\pdfoutput=1

\documentclass[11pt]{article}

\usepackage[]{acl}

\usepackage{times}
\usepackage{latexsym}

\usepackage[T1]{fontenc}

\usepackage[utf8]{inputenc}

\usepackage[patch=none]{microtype}

\usepackage{inconsolata}

\usepackage{graphicx}
\usepackage{xcolor}
\usepackage{multirow}
\usepackage{booktabs}
\usepackage{longtable}
\usepackage{cprotect}
\usepackage{hyperref}
\usepackage{linguex}

\usepackage{amsmath,amssymb}
\usepackage{enumitem}

\newcommand*{\textbox}[1]{\framebox{#1}}

\definecolor{trueblue}{rgb}{0.0, 0.45, 0.81}

\definecolor{NavyBlue}{HTML}{006EB8}
\definecolor{Red}{HTML}{ED1B23}
\definecolor{Orange}{HTML}{F58137}
\definecolor{Green}{HTML}{00A64F}

\title{Integrating gender inclusivity into large language models\\via instruction tuning}

\author{Alina Wróblewska \and Bartosz Żuk \\
  Institute of Computer Science \\
  Polish Academy of Sciences, Warsaw, Poland \\
  \texttt{\{alina, b.zuk\}@ipipan.waw.pl}\\\\
  \includegraphics[width=4mm]{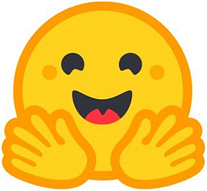} \href{https://huggingface.co/datasets/ipipan/IPIS}{Dataset} \hspace{5mm} \includegraphics[width=4mm]{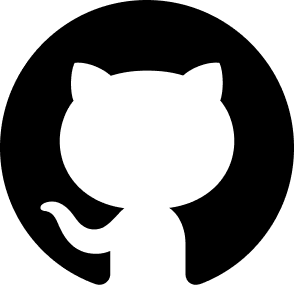} \href{https://github.com/ipipan/IPIS/}{Code} \hspace{5mm} \includegraphics[width=4mm]{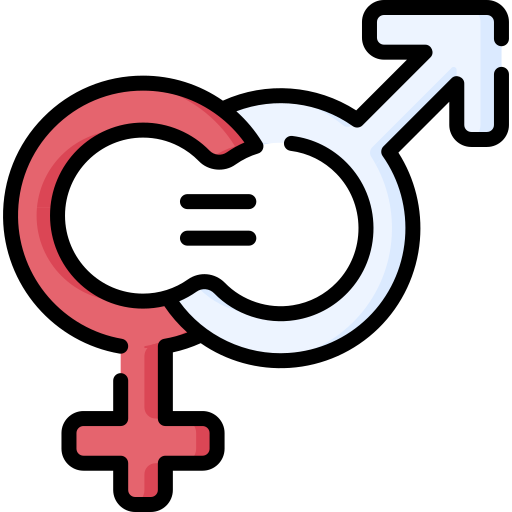} \href{https://huggingface.co/collections/ipipan/ipis-models}{Models}}
\begin{document}
\maketitle
\begin{abstract}
Imagine a language with masculine, feminine, and neuter grammatical genders, yet, historically and politically entrenched norms have established masculine forms as the default for referring not only to men but also to women and mixed-gender groups. This is the~reality of contemporary Polish and several other Slavic and Romance languages. As a result, large language models (LLMs) trained on biased corpora tend to reproduce androcentric patterns, generating gender-imbalanced outputs and impeding efforts towards gender-inclusive language.\\
In this study, we address the problem of default androcentrism in LLMs by adapting them to gender-inclusive behaviour in Polish. We introduce the Inclusive Polish Instruction Set (IPIS), a~collection of human-authored, gender-inclusive instructions for proofreading and Polish$\leftrightarrow$English translation. Grounded in linguistic theory, we design a system prompt with explicit guidelines for gender-inclusive Polish. We tune both multilingual and Polish-specific LLMs using IPIS and evaluate their output in terms of gender inclusivity.\\
Our results show that gender inclusivity can be effectively learnt through instruction tuning across both investigated tasks: gender-inclusive proofreading and gender-sensitive translation. Our work thus provides a practical approach to mitigating gender bias in Polish and contributes to broader efforts to align LLMs with socially responsible linguistic practices.
\end{abstract}

\section{Introduction}
\label{sec:intro}
\begin{figure*}
\centering
\includegraphics[width=\textwidth]{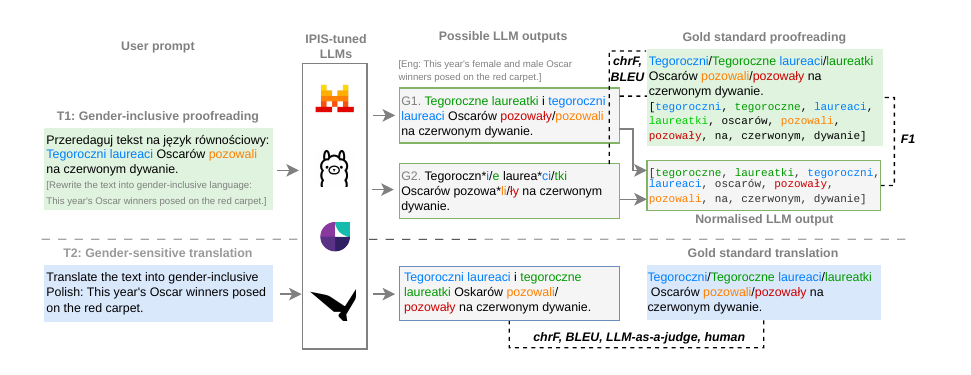}
    \caption{\label{fig:flowchart} Flowchart of inference and evaluation workflows for IPIS-tuned LLMs for \textsl{gender-inclusive proofreading} (top) and Polish$\leftrightarrow$English \textsl{gender-sensitive translation} (bottom; English$\rightarrow$Polish direction shown). Explanations: \textcolor{NavyBlue}{masculine nouns and adjectives}, \textcolor{Orange}{masculine verbs}, \textcolor{Green}{feminine nouns and adjectives} and \textcolor{Red}{feminine verbs}.}
\end{figure*}

Large language models (LLMs) are trained on extensive datasets sourced primarily from the internet, reflecting dominant usage patterns and, consequently, socially embedded biases and stereotypes \cite{doi:10.1126/science.aal4230, 10.1145/3442188.3445922}. LLMs are highly effective in internalising and replicating harmful patterns \cite{lucy-bamman-2021-gender,fleisig-etal-2023-fairprism}. In this study, we focus on one such phenomenon: \textit{gender bias}. In English, gender bias is mainly lexical and contextual \cite{stanczak2021surveygenderbiasnatural}, and LLMs tend to reflect a~gender-binary orientation \cite{ovalle-etal-2024-tokenization}. 
In Polish, gender bias is structural and grammatical -- masculine nouns are often used as a generic default to refer to men, women, and non-binary people \cite{KoniuszaniecBlaszkowska+2003+259+285,Stahlberg2007RepresentationOT,gygax:2019}. Consequently, LLMs trained on Polish data acquire masculine-biased grammatical patterns and tend to favour a~masculine orientation \cite{wro:inni:2025}.

The dominance of masculine expressions over feminine ones 
is a~form of sociolinguistic gender discrimination \cite{geg:2016,gnl:2018}. Recognising its harmful effects, the Council of Europe encourages member states to eliminate sexist language and promote gender-equal linguistic practices \cite{cm:rec:2019}.

Gender bias and inclusiveness are central concerns in modern natural language processing (NLP) and natural language generation (NLG), as reflected in dedicated workshops: GeBNLP \citep{gebnlp-ws-2025-1,gebnlp-2026-1}
and GITT \cite{gitt-2025-1,gitt-2026-1}. With the growing significance of gender-inclusive NLP, particularly in machine translation \cite[MT; ][]{menis-mastromichalakis-etal-2025-assumed,savoldi-etal-2025-mind} and gender-neutral rewriting \cite{vanmassenhove-etal-2021-neutral,sun2021theythemtheirsrewriting}, a key goal is to develop LLMs that reduce bias and use inclusive language.

This study takes an initial step toward this goal for Polish. We investigate whether instruction tuning can produce gender-inclusive LLMs (Figure \ref{fig:flowchart}) and evaluate them in Polish proofreading and PL$\leftrightarrow$EN translation. To our knowledge, this is the first instruction-tuning approach to gender-inclusive LLM adaptation for a Slavic language and the first to use a fully human-annotated dataset.

Our contributions are threefold: 
\textbf{(1)} We release the Inclusive Polish Instruction Set (\href{https://huggingface.co/datasets/ipipan/IPIS}{IPIS}) (Section \ref{sec:ipis_dataset}), the dataset for gender-inclusive proofreading and gender-sensitive translation with expert-annotated gender-fair Polish texts, 
\textbf{(2)} We design a system prompt to guide LLMs in generating gender-inclusive language (Section \ref{sec:prompts}). \textbf{(3)}~We evaluate how different instruction-tuning setups affect gender inclusivity in Polish proofreading and Polish$\leftrightarrow$English translation (Sections \ref{sec:evaluation_methodology}, \ref{sec:results_main}), and release best-performing \href{https://huggingface.co/collections/ipipan/ipis-models}{IPIS-tuned LLMs} and \href{https://github.com/ipipan/IPIS/}{code}.

\section{Proposed approach}
\label{sec:approach}
This study examines instruction-tuning as a stra\-te\-gy for developing gender-inclusive LLMs for Polish.

\subsection{Polish -- an androcentric language}
\label{sec:polish}

Polish is a \textit{grammatical gender language} \cite{REALI_ESAULOVA_VON_STOCKHAUSEN_2015} in which all nouns are marked for \textit{grammatical gender}:\footnote{\textit{Grammatical gender} is a category that classifies nouns into declension groups and governs agreements in a sentence.} \textit{śliwka} [plum] is feminine, \textit{pomidor} [tomato] is masculine and \textit{jabłko} [apple] is neutral. Polish verbs and adjectives show grammatical agreement with nouns. Some \textit{personal nouns} \cite{BACKER2012253} are realised as gendered dyads, e.g. \textit{prawnik}/\textit{prawniczka} [lawyer$_{\text{masc/fem}}$]. Feminine personal nouns refer to women, whereas masculine nouns can refer to men, mixed or unspecified groups, and even women (\textit{generic masculine}), e.g. \textit{niemiecka polityk, Ursula von der Leyen} 
[German$_{\text{fem}}$ politician$_{\text{masc}}$, Ursula$_{\text{fem}}$ von der Leyen].

Although its grammatical system allows marking \textit{referential gender},\footnote{\textit{Referential gender} refers to the sex (female or male) or gender identity of a living referent.} Polish is heavily masculine-oriented, with a dominance of masculine expression and reinforcement of gender bias. Adapting Polish to reflect referential genders is a challenging and time-consuming process. Its goal is \textit{gender-inclusive} (or \textit{gender-fair}) language with all genders explicitly marked \cite{amrhein-etal-2023-exploiting}. Gender-inclusive Polish may also use expressions that are neutral with respect to referential gender (e.g. \textit{osoba z niepełnosprawnością} [person with a disability]). However, such expressions do not eliminate grammatical gender marking, i.e. \textit{osoba} [person$_{\text{fem}}$] does not specify the gender of an individual being referred to, but it is grammatically feminine. We therefore avoid the term \textit{gender-neutral} language, which generally refers to languages that avoid explicit marking gender.

As LLMs trained on Polish corpora reproduce dominant androcentric patterns \cite{wro:inni:2025}, we investigate whether targeted instruction tuning can enable gender-sensitive language generation. We focus on well-established feminine forms, masculine–feminine dyads, and gender-neutral expressions, leaving room for future inclusion of non-binary forms (not yet standardised). Our goal is to make gender inclusivity a~core modelling principle, enabling consistent and context-appropriate generation of gender-fair Polish.

\subsection{Gender-inclusive instruction tuning}
\label{sec:tuning}

Instruction tuning is a fine-tuning technique that improves the versatility and controllability of LLMs on a wide range of tasks \cite{zhang2024instructiontuninglargelanguage}. It consists in training LLMs on instruction--response pairs, following a supervised learning paradigm.

We use instruction tuning to adapt pre-trained LLMs to gender-inclusive language use by exposing them to a set of 
instructions with gender-inclusive responses (see Section \ref{sec:ipis_dataset}). For this, we employ \textit{low-rank adaptation} \cite[LoRA,][]{hu2021loralowrankadaptationlarge}, a parameter-efficient fine-tuning method. LoRA updates only a small subset of parameters by injecting low-rank decomposition matrices into transformer layers while keeping the original weights frozen. This reduces computational and memory requirements, enables controlled gender-inclusive adaptation without full retraining, and preserves the model's general linguistic competence. This well-established method also supports reproducibility. 

\subsection{Proposed gender-inclusive tasks}
\label{sec:tasks}

In the proposed approach, we adapt LLMs to perform two tasks: proofreading and translation.

\paragraph{Gender-inclusive proofreading} This task consists in transforming the conventional Polish text into its gender-inclusive version. It goes beyond surface-level paraphrasing, as models must reason about referential gender and apply linguistically informed transformations. Specifically, LLMs must:
\begin{itemize}
    \item distinguish personal nouns from non-personal ones (including collective nouns, e.g. \textit{komitet} [committee$_{\text{masc}}$]); \item modify generic masculine nouns, but preserve masculine forms that refer genuinely to men or male-only groups;
    \item preserve epicene nouns that are gender-neutral despite their grammatical gender (e.g. \textit{rodzic} [parent$_{\text{masc}}$]) and inherently gender-specific nouns (e.g. \textit{wujek} [uncle$_{\text{masc}}$]).
\end{itemize}
LLMs must generate linguistically and socially accepted gender-inclusive forms and correctly adjust all grammatical agreement (i.e. adjective, verb, and pronoun forms), including implicit references in pro-drop contexts.

\paragraph{Gender-sensitive translation} This task focuses on bidirectional translation between English and gender-inclusive Polish. In PL$\rightarrow$EN, the goal is to generate fluent, non-redundant English translations from Polish texts containing gendered dyads of nouns, verbs, adjectives. In EN$\rightarrow$PL, the goal is to translate gender-unmarked expressions into contextually appropriate Polish using gender-specific expression, gendered dyads, or neutral forms (see examples of translation challenges in Appendix~\ref{sec:appendixC2}).

\section{Inclusive Polish Instruction Set (IPIS)}
\label{sec:ipis_dataset}
The IPIS dataset is a collection of instructions to improve gender sensitivity and inclusiveness of Polish LLMs.
\paragraph{Gender-inclusive annotation} The IPIS instructions are based on human-annotated texts from various open-access sources representing diverse genres: legal acts, institutional texts, narratives, news, transcripts, etc. (all data sources are listed in Appendix \ref{sec:apx:ipis_content}). Originally androcentric texts are manually rewritten into inclusive versions by two annotators and reviewed by a third, producing a parallel androcentric--to--gender-inclusive Polish corpus. 
Annotators apply feminisation -- gendered dyads (e.g. \textit{szli i szły}, \textit{szli/szły}, \textit{sz*li/ły} [they walked$_{\textsl{masc/fem}}$]) and feminine forms (e.g. \textit{kierowniczka} [manageress]), neutralisation -- impersonal expression (e.g. \textit{powiedziano} [it was said]) or collective nouns (e.g. \textit{kadra} [staff]), or combinations of these strategies, and produce fluent, linguistically accurate and contextually appropriate gender-inclusive texts (annotation details and examples are in Appendices \ref{sec:apx:ipis_details} and \ref{sec:apx:strategies_illustrations}).

Even when restricted to valid expressions, gender-inclusive proofreading remains inherently open-ended, given the multiple possible notations and orderings of gender dyads (see examples in Appendix \ref{sec:appendixC}). Consequently, standard inter-annotation agreement measures are unsuitable for this task.
Instead, we measure agreement on where changes are introduced, reaching 93.5\%/94.3\% at the text level and 86.7\%/89.6\% at the token level for the two independent 
annotations (see annotation statistics in Appendix \ref{sec:apx:ipis_statistical_summary}).

\paragraph{IPIS dataset}
Each IPIS-proofreading sample consists of:

\begin{enumerate}
    \setlength\itemsep{-.2em}
    \item \textbf{user prompt} (\texttt{prompt}) -- a 1-2 sentence instruction that specifies a given task and resembles a real-world request (e.g. \textit{Rewrite...}),
    \item \textbf{input text passage} (\texttt{source}) -- an androcentric passage requiring proofreading,
    \item \textbf{desired output} (\texttt{target}) -- an expected gender-inclusive response based on \texttt{prompt} and \texttt{source}, which is used as ground truth for evaluating and optimising LLM's predictions.
\end{enumerate}

\noindent
IPIS-translation instances additionally include \texttt{(prompt|source|target)\_language} with values restricted to PL (Polish) or EN (English). The IPIS instruction examples are shown in Figure \ref{fig:flowchart} and Appendix  \ref{sec:apx:ipis_instances}.

The monolingual IPIS-proofreading dataset contains 5,278 instances in \textsl{test}, 2,732 in \textsl{dev}, and 23,532 in \textsl{train}. The bilingual IPIS-translation dataset is provided in two versions. The \textbf{human-written} version (human-authored on both sides) contains 1,728/304/456 \textsl{train}/\textsl{dev}/\textsl{test} pairs. The \textbf{mixed-generated} version pairs all human-written gender-inclusive IPIS-proofreading texts with automatic English translations. To mitigate the risk of MT inconsistencies -- given that LLMs are trained on androcentric Polish corpora -- we translate the IPIS-proofreading source passages rather than gender-inclusive target ones. Translations are generated using Gemma-3-27B \cite{gemmateam2025gemma3technicalreport}, the best-performing model on standard PL$\rightarrow$EN translation (BLEU: $53.81$; see the lower part of Appendix Table \ref{tab:translation_D} for performance results on standard translation).

The IPIS-proofreading splits (and, consequently, mixed-generated splits) are balanced for gender-inclusive transformation types. Gendered dyads account for nearly $78\%$ of all changes, including coordination ($\approx\!38\%$), slash notation ($\approx\!15\%$), gender-star forms ($\approx\!25\%$). The remaining $\approx\!22\%$ consists of neutral and feminine expressions (see Appendix Table \ref{tab:annotation_strategies} for the full distribution). IPIS is used for LLM tuning and evaluation.

\section{Experimental setup}
\label{sec:experiments}
\paragraph{Tested LLMs}
\label{sec:tested_llms}
Our study investigates whether LLMs can be effectively fine-tuned on IPIS, ensuring socially responsible language generation in Polish. 
We conduct a series of experiments using open-source LLMs based on the transformer architecture \cite{NIPS2017_3f5ee243}. We focus on instruction-tuned, small- and medium-sized versions of both \textbf{multilingual LLMs}: \textsl{Llama-8B}, \textsl{Mistral-7B}, and \textsl{Mistral-Nemo}, and \textbf{Polish-specific models}: \textsl{Bielik-7B}, \textsl{Bielik-11B}, \textsl{Llama-PLLuM-8B}, and \textsl{PLLuM-12B} (see Appendix \ref{sec:appendix7a} for model details).

\paragraph{System prompt}
\label{sec:prompts}
We study the effect of including a system prompt with gender-inclusive guidelines during training and evaluation. The system prompt, grounded in the theoretical framework developed by \citet{wro:inni:2025}, is originally written in Polish and translated into English (Appendix \ref{sec:appendixA}).

Using persona prompting \cite{xu2023expertpromptinginstructinglargelanguage,kong-etal-2024-better}, we assign LLMs the role of an editor or translator. They are instructed to follow an~algorithm and proofread/translate texts according to the gender-inclusive principles specified for a particular genre. 
The system prompt defines exclusionary expressions, provides inclusive alternatives, and specifies agreement rules to ensure grammatically correct and gender-inclusive outputs.

\paragraph{Implementation details}
\label{sec:implementation_details}

LLMs are instruction-tuned using DeepSpeed ZeRO, Flash Attention and HuggingFace’s ecosystem (\textsl{transformers}, \textsl{trl}, \textsl{peft} and \textsl{datasets} libraries). DeepSpeed is a library designed for compute and memory optimisation in distributed training setups. HuggingFace’s  \textsl{datasets} and \textsl{transformers} libraries facilitate the loading and usage of the IPIS dataset and selected LLMs while \textsl{trl} provides a dedicated trainer for supervised fine-tuning (SFT). LoRA adapters are configured using HuggingFace’s \textsl{peft} library. Flash Attention enhances the efficiency of the core attention mechanism in transformer-based models. Details on LoRA configuration, SFT hyperparameters, and the libraries used are in Appendix \ref{sec:appendixB}.

During instruction tuning, we compute the cross-entropy loss for both prompt and completion tokens (default setting in \textsl{trl.SFTTrainer}), meaning that the model also learns the structure of the input requests. The prompt and completion losses are minimised on the \textsl{train} set while being tracked on the \textsl{dev} set.

\section{Evaluation methodology}
\label{sec:evaluation_methodology}
\paragraph{Evaluation metrics}
For gender-inclusive proofreading, we evaluate normalised LLM outputs against gold-standard test instances using precision, recall, and F$_1$-measure. Normalisation includes lowercasing, removing punctuation and conjunctions, and transforming all gender-inclusive expressions into full masculine and feminine forms. Punctuation and conjunctions are identified using the Lambo tokeniser
\cite{LAMBO} and COMBO-NLP \cite{combo_nlp_2026,klimaszewski-wroblewska-2021-combo-state}. Figure \ref{fig:flowchart} illustrates the normalisation of possible LLM outcomes -- G1 with gendered dyads and G2 with gender stars and slashes. The normalisation procedure is presented in detail with examples in Appendix \ref{sec:appendixC}. 

The same normalisation was applied to the gender-inclusive reference texts and source texts (before proofreading). This allows the normalised LLM output to be directly compared with both the normalised gold-standard and the normalised source tokens. Since normalised tokens are treated as bags of words rather than ordered sequences, word order is irrelevant. We can thus calculate \textsl{true positives} (tokens introduced by the model that are present in the gold-standard but absent from the source), \textsl{true negatives} (tokens present in all three lists), \textsl{false positives} (tokens introduced by the model that are not present in the gold-standard, including unnecessary changes and out-of-vocabulary tokens) and \textsl{false negatives} (tokens required by the gold-standard but missing from the model output). IPIS may contain two different but equally valid gender-inclusive references for the same source text, proposed independently by two annotators, and accepted by the superannotator. In such cases, we evaluate the model output against both references and retain the results obtained with the better-matching reference. 

Additionally, we measure similarity between LLM-proofread and reference gender-inclusive passages in their original, not-normalised form, using the machine translation metrics -- chrF \cite{popovic-2015-chrf} and \textsc{bleu} \cite{papineni-etal-2002-bleu}. Since LLMs should improve inclusivity while preserving structure and content, and not hallucinate, these metrics are complementary adequacy indicators.

\textsc{bleu} and chrF are also used to evaluate bidirectional translations between English and gender-inclusive Polish. Since these metrics may not capture gender-inclusive changes, particularly in EN$\rightarrow$PL, we also perform LLM-as-a-judge evaluation and compare its scores with human ratings under the same scoring protocol. Following G-Eval \cite{liu-etal-2023-g}, the judge evaluates translation quality (\textsl{completeness}, \textsl{clarity and logical coherence}, \textsl{structural fidelity}, \textsl{linguistic correctness}) and \textsl{gender inclusivity} on a 1--5 scale (see the evaluation prompt in Appendix Figure~\ref{fig:llm-as-a-judge-prompt}). The general-capability evaluation provides an additional sanity check to verify that the translation results are not due to model degradation or methodological issues.

\paragraph{Baseline scenarios}
We evaluate the impact of instruction tuning and system-prompt guidance using several baseline configurations. The main model is an IPIS-tuned LLM (\textsl{tuned}), which can additionally be provided with the system prompt (\textsl{tuned}-\{\textsl{pl}|\textsl{en}\}). All IPIS-tuned LLMs are compared with their off-the-shelf counterparts in a~zero-shot setting (\textsl{default}), optionally augmented with the system prompt (\textsl{default}-\{\textsl{pl}|\textsl{en}\}). Further baselines include few-shot variants of default models, evaluated without (\textsl{fewshot}) or with the system prompt (\textsl{fewshot}-\{\textsl{pl}|\textsl{en}\}). This experimental setting enables a~systematic comparison by isolating the effects of instruction tuning, system-prompt guidance and in-context learning.

\section{Results and discussion}
\label{sec:results_main}
\subsection{Gender-inclusive proofreading evaluation}
\label{sec:results}

\paragraph{Impact of the LLM leading language}

Polish-specific models (see Section \ref{sec:tested_llms}), obtained by further tuning multilingual LLMs on large Polish corpora, are intended to improve performance on tasks requiring a deeper understanding of Polish. However, in both zero-shot (\textsl{default}) and few-shot (\textsl{fewshot}) settings, they show no improvement over their multilingual predecessors, achieving F$_1$ scores close to zero (see Appendix Table \ref{tab:proofreading_D} for all proofreading results). This finding, while disappointing, is expected, as the additional Polish training data predominantly contain non-inclusive language.

\begin{table}[!ht]
{\small
\renewcommand\tabcolsep{6.7pt}
\begin{tabular}{l||rrr|rr}
\toprule
\textbf{LLM} & \textbf{Prec} & \textbf{Rec} & \textbf{ F$_1$} & \textbf{\textsc{bleu}} & \textbf{chrF}\\
 \midrule
 \multicolumn{6}{c}{Llama-PLLuM-8B}\\
 \midrule
 \textsl{default}   & 0.25 & 0.30 & 0.27 & 32.04 & 57.13 \\
 \textsl{fewshot}   & 0.54 & 0.64 & 0.59 & 37.51 & 52.38\\
 \textsl{tuned}     & 69.72 & 52.79 & 60.08 & 94.28 & 97.64 \\
  \midrule
  \multicolumn{6}{c}{Bielik-11B}\\
  \midrule
 \textsl{default}   & 0.40 & 0.74 & 0.52 & 39.12 & 66.54\\
 \textsl{fewshot}   & 1.20 & 5.10 & 1.94 & 52.91 & 80.23\\
 \textsl{tuned}     & \textbf{72.68} &  \textbf{65.09} &  \textbf{68.68} & \textbf{95.22} &  \textbf{97.99}\\
 \midrule
 \multicolumn{6}{c}{Mistral-Nemo-12B}\\
 \midrule
 \textsl{default}   & 0.42 & 0.47 & 0.44 & 62.98 & 78.33\\
 \textsl{fewshot}   & 0.84 & 1.90 & 1.16 &  68.43 &  84.95\\
 \textsl{tuned}     & 64.82 & 56.24 & 60.23 & 94.45 & 97.72\\
 \bottomrule
\end{tabular}}
\caption{\label{tab:token_language} \textbf{Leading language} and \textbf{LLM size} factors. Performance of the multilingual Mistral-Nemo-12B model and the Polish-specific Llama-PLLuM-8B and Bielik-11B models in their \textsl{default}, \textsl{fewshot} and IPIS-\textsl{tuned} versions on the gender-inclusive proofreading task.}
\end{table}

Table \ref{tab:token_language} reports the scores of the best-performing IPIS-tuned models -- Mistral-Nemo-12B (multilingual) and Bielik-11B (Polish-specific), which substantially outperform their baseline configurations (\textsl{default} and \textsl{fewshot}). The leading language of an IPIS-tuned LLM appears to influence its effectiveness in gender-inclusive proofreading, with the Polish-specific Bielik-11B achieving a higher F$_1$ score ($68.68$) than Mistral-Nemo (F$_1= 60.23$).

In terms of \textsc{bleu} and chrF similarity scores, the IPIS-tuned models achieve similarly high scores, confirming the effectiveness of instruction tuning in improving gender inclusiveness of LLMs. However, the \textsl{default} and \textsl{fewshot} baselines show a~clear advantage of Mistral-Nemo-12B over Bielik-11B (for the \textsl{default} baselines: \textsc{bleu} $= 62.98$, chrF $= 78.33$ vs. \textsc{bleu} $= 39.12$, chrF $= 77.54$, respectively), suggesting that large-scale multilingual pretraining offers benefits beyond language-specific adaptation. Moreover, the \textsl{fewshot} baselines generally outperform their \textsl{default} counterparts, particularly for Bielik-11B (\textsl{default} chrF $=66.54$ vs. \textsl{fewshot} chrF $=80.23$), indicating that in-context examples improve proofreading performance, although they remain insufficient to match the performance of the IPIS-tuned models.

\paragraph{Impact of the LLM size}
As expected, larger IPIS-tuned models generally outperform smaller ones, as reflected in their average F$_1$ scores: $65.23$ vs. $55.93$ (see Appendix Table \ref{tab:proofreading_D} for all results). Table \ref{tab:token_language} compares the best small and medium IPIS-tuned LLMs -- Llama-PLLuM-8B and Bielik-11B, respectively.

\begin{table}[!ht]
{\small
\renewcommand\tabcolsep{6pt}
\begin{tabular}{l||rrr|rr}
\toprule
\textbf{LLM}  & \textbf{Prec} & \textbf{Rec} & \textbf{ F$_1$} & \textbf{\textsc{bleu}} & \textbf{chrF}\\
 \midrule
\textsl{default}     & 0.40 & 0.74 & 0.52 & 39.12 & 66.54\\
 \textsl{default-pl} & 1.57 & 10.46 & 2.74 & \underline{56.56} & 83.93\\
 \textsl{default-en} & \underline{1.73} & \underline{15.34} & \underline{3.11} & 55.79 & \underline{84.72}\\
 \midrule
\textsl{fewshot}     & 1.20 & 5.10 & 1.94 & 52.91 & 80.23\\
 \textsl{fewshot-pl} & \underline{1.43} & \underline{13.10} & \underline{2.58} & \underline{54.92} & \underline{84.01}\\
 \textsl{fewshot-en} & 1.41 & 9.71 & 2.47 & 53.69 & 82.93\\
  \midrule
  \midrule
\textsl{tuned}       & \textbf{72.68} &  \textbf{65.09} &  \textbf{68.68} & \textbf{95.22} &  \textbf{97.99}\\
 \textsl{tuned-pl}   & 32.09 & 57.58 & 41.22 & 91.82 & 96.93\\
 \textsl{tuned-en}   & 58.67 & 59.24 & 58.95 & 94.82 & 97.61\\
 \bottomrule
\end{tabular}}
\caption{\label{tab:token_system_prompt} \textbf{System prompt} factor. Performance of the Polish-specific Bielik-11B model in its \textsl{default}, \textsl{fewshot} and IPIS-\textsl{tuned} versions, possibly with a system prompt (\textsl{pl} or \textsl{en}) on the gender-inclusive proofreading task.}
\end{table}

\paragraph{Impact of the system prompt}
All models are trained with a system prompt in Polish (\textsl{pl}) or English (\textsl{en}), and without it. Table \ref{tab:token_system_prompt} reports the scores for the best-performing IPIS-\textsl{tuned} Bielik-11B model and its \textsl{default} and \textsl{fewshot} variants. The IPIS-\textsl{tuned} Bielik-11B trained without any system prompt significantly outperforms all its baseline models, and it also achieves superior proofreading performance compared to its IPIS-tuned alternatives guided by system prompts. This pattern generalises across evaluated LLMs: the system prompts generally benefit the baseline models but not the IPIS-tuned ones (see Appendix Table~\ref{tab:proofreading_D} for all results). A plausible explanation is that instruction tuning already internalises the desired gender-inclusive behaviour, making the system prompt largely redundant.

The language of the system prompt (Polish or English) does not have a significant impact on performance. Although the Polish prompt tends to be marginally more effective for baseline models and the English prompt for IPIS-tuned models, these differences are negligible and insignificant.

\paragraph{Key findings} IPIS-tuned LLMs substantially outperform baseline models. Given that IPIS-tuned Polish-specific models are more inclusive than their multilingual variants, we infer that language-specific models are better suited for instruction-level optimisation in Polish. Moreover, IPIS-tuned LLMs consistently achieve higher precision than recall. This suggests that they introduce gender-inclusive modifications primarily when they are highly confident, minimising false positives. Although this cautious strategy improves precision, it also results in the omission of required gender-inclusive edits. Both  \textsc{bleu} (word \textit{n}-gram-based) and chrF (character \textit{n}-gram-based) scores are comparable and exceed $90$ across all IPIS-tuned models, which means that proofread texts are of high quality, with minimal out-of-vocabulary words, typos, morphosyntactic errors, or lexical mismatches. In general, these findings show that IPIS-tuned LLMs represent a significant step toward the development of truly gender-inclusive LLMs for Polish.
 
\subsection{Evaluation of gender-sensitive translation}

\paragraph{Preliminary experiments} 
We fine-tune and evaluate LLMs on the small \textbf{human-written} version of the IPIS-translation dataset. Medium-sized models outperform smaller ones (see Appendix Table \ref{tab:translation_D} for all results), with Bielik-11B performing best overall. In this preliminary setting, benefits of IPIS tuning are mixed (see \textsl{hum-tuned} rows in Table \ref{tab:translation_short}). Although the IPIS-tuned Bielik-11B performs best for PL$\rightarrow$EN translation (\textsc{bleu} $= 57.55$; chrF $= 78.03$), its few-shot version yields the best EN$\rightarrow$PL results (\textsc{bleu} $= 43.02$; chrF $= 72.46$).

These results suggest that translating gender-inclusive Polish into English is relatively easy for LLMs, and even limited IPIS tuning improves Bielik-11B over the baselines. EN$\rightarrow$PL, in turn, remains more challenging. The IPIS-tuned Bielik-11B underperforms its few-shot counterpart, indicating that the available human-written IPIS data are not yet sufficiently large (2488 instances in total) and diverse (European legal acts) to support high-quality gender-inclusive Polish translation.

 \begin{table}[!ht]
\renewcommand\tabcolsep{7pt}
{\small
\centering
\begin{tabular}{l||c|c||c|c}%
    \toprule
        \multirow{2}{*}{\textbf{LLM}} & \multicolumn{2}{c||}{\textbf{Polish$\rightarrow$English}} & \multicolumn{2}{c}{\textbf{English$\rightarrow$Polish}}\\
         \cmidrule(lr){2-3}  \cmidrule(lr){4-5}
        & \textbf{\textsc{bleu}} & \textbf{chrF}  & 
         \textbf{\textsc{bleu}} & \textbf{chrF} \\
    \midrule
    \multicolumn{5}{c}{\textbf{human-written} IPIS-translation-test}\\
\midrule
\textsl{default} & 47.60 & 73.39 &  41.49 & 71.78  \\
\textsl{default-pl} & 46.78 & 73.08    & 35.80& 69.77 \\
\textsl{default-en} & 47.99 & 73.76    & 32.70 & 68.65  \\
\midrule
\textsl{fewshot} & 50.01 & 73.93    & 38.63 & 68.78 \\
\textsl{fewshot-pl} & 49.38 & 73.84  & 37.81 & 69.92  \\
\textsl{fewshot-en} & 48.33 & 73.43  & \textbf{43.02} & \textbf{72.46}  \\
\midrule
\textsl{hum-tuned} & 55.19 & 75.80  & 34.26 & 55.08  \\
\textsl{hum-tuned-pl}  & 56.70 & 76.93   & 31.70 & 58.36  \\
\textsl{hum-tuned-en} & \textbf{57.55} & \textbf{78.03}  & 28.71 & 60.97 \\
\midrule
\textsl{mix-tuned} & 57.27 & 77.90  & 34.29 & 63.93 \\
\textsl{mix-tuned-pl} & 54.67 & 75.82  & 29.84 & 61.16  \\
\textsl{mix-tuned-en} & 53.76 & 75.08  & 33.26 & 63.86 \\
\midrule
\textsl{nllb-default} & 45.26 & 65.97 & 32.43 & 63.31 \\
\midrule
\multicolumn{5}{c}{\textbf{mixed-generated} IPIS-translation-test}\\
\midrule
\textsl{hum-tuned} & 35.58 & 63.78 & 21.23 & 53.93\\
\textsl{hum-tuned-pl} & 33.02 & 61.54 & 26.08 & 56.64\\
\textsl{hum-tuned-en} & 34.24 & 63.79 & 25.91 & 57.14 \\
\midrule
\textsl{mix-tuned} & \textbf{60.35} & \textbf{77.69}  & \textbf{46.57} & \textbf{70.60} \\
\textsl{mix-tuned-pl} & 55.89 & 75.04 & 42.81 & 68.54 \\
\textsl{mix-tuned-en} & 54.94 & 75.16 & 43.12 & 68.69  \\
\midrule
 \textsl{nllb-default} & 25.90 & 48.77 & 21.68 & 50.71 \\
\bottomrule
\end{tabular}}
\caption{\label{tab:translation_short} Performance of Bielik-11B on the \textbf{gender-sensitive translation} task across \textsl{default}, \textsl{fewshot} and IPIS-\textsl{tuned} settings, with Polish user prompts and optional system prompts (\textsl{-pl}, \textsl{-en}). The model is fine-tuned on either human-written (\textsl{hum-}) or mixed-generated (\textsl{mix-}) data; corresponding test sets are used for evaluation (top vs. bottom). Results from the NLLB-200-distilled-1.3B model are reported for comparison.
}
 \end{table}

\paragraph{Follow-up experiments} During instruction tuning, an LLM updates its parameters to reflect task-specific examples; however, instruction tuning on small datasets can lead to overfitting, limiting generalisation \cite{shi2024instructiontuninglossinstructions}. A common mitigation strategy is data augmentation. We thus tune Bielik-11B on the full \textbf{mixed-generated} IPIS-translation set (see \textsl{mix-tuned} rows in Table \ref{tab:translation_short}). The \textsl{mix-tuned} model without a system prompt consistently outperforms the \textsl{hum-tuned} counterpart, achieving higher chrF scores in both translation directions ($77.9$ vs. $75.8$ for PL$\rightarrow$EN, and $63.93$ vs. $55.08$ for EN$\rightarrow$PL), when tested on the human-written test set. Moreover, when evaluated on the mixed-generated test set (see the lower part of Table \ref{tab:translation_short}), the \textsl{mix-tuned} model substantially outperforms the \textsl{hum-tuned} variant, indicating improved robustness compared to the preliminary translation setting. In contrast, the \textsl{hum-tuned} model performs poorly when evaluated on the mixed-generated test set. This supports our preliminary assumption that the size and diversity of the training data are crucial to robust performance.

\paragraph{Language impact} The language of user queries (prompts) does not significantly impact translation quality, as the scores for Polish and English user prompts within the same translation direction are largely comparable. The use of a system prompt -- especially the English one -- improves performance in some cases (e.g. in the \textsl{hum-tuned} PL$\rightarrow$EN setting), but generally provides marginal benefits and may even degrade performance. 

\paragraph{SOTA comparison}
To contextualise our results, we compare them with the state-of-the-art translation model -- \textsl{NLLB-200-distilled-1.3B} \cite{nllbteam2022languageleftbehindscaling} (see the \textsl{nllb-default} rows in Table \ref{tab:translation_short}). All Bielik-11B configurations -- both baselines and IPIS-tuned -- significantly outperform NLLB in both translation directions and on both test sets. NLLB performs particularly poorly on the mixed-generated test set, indicating limited robustness to gender-inclusive language and highlighting both the effectiveness and necessity of our IPIS-based adaptation.

\paragraph{LLM-as-a-judge and human evaluation}
Although all IPIS-tuned Bielik-11B models outperform the SOTA NLLB model, \textsc{bleu} and chrF do not capture gender-inclusive changes, especially in EN$\rightarrow$PL translation. We therefore complement automatic evaluation with LLM-as-a-judge and human evaluation on 114 human-written IPIS-translation instructions. We use \textsl{Qwen3-30B-A3B-Thinking-2507} \cite{qwen3technicalreport} to judge Polish translations generated by Bielik-11B.

Across all criteria, the LLM-judge assigns lower scores than the human evaluator, suggesting its stricter scoring behaviour (see Appendix \ref{sec:appendix_llm_human} and its Table \ref{tab:llm-as-a-judge-table} for a detailed comparison of LLM-as-a-judge and human evaluations of EN$\rightarrow$PL translations). For MT-related criteria, especially adequacy-related aspects, both evaluators assign higher scores to the highest-\textsc{bleu} translations than to the lowest-\textsc{bleu}  ones (for \textsl{completeness}, LLM: $2.632$ vs. $1.281$, Hum: $4.053$ vs. $1.333$). In contrast, the difference between the highest- and lowest-\\textsc{bleu} subsets is significantly smaller for \textsl{gender inclusivity} (LLM: $3.263$ vs. $2.947$, Hum: $4.246$ vs. $3.579$). This indicates that gender-inclusive language choices are only weakly associated with \textsc{bleu}-based lexical similarity, and $n$-gram metrics may fail to capture non-standard, socially relevant linguistic properties. At the same time, LLM and human inclusivity scores show a moderate, statistically significant correlation (Pearson $r = 0.517$; Spearman 
$\rho = 0.539$), suggesting that LLM-based evaluation approximates human judgments.

These results suggest that \textsl{gender inclusivity} behaves as an independent dimension of translation quality, i.e. inclusive expressions are still present in low-quality translations (see additional analysis in Appendix \ref{sec:apx:modifications}). As LLM can generate high- and low-quality translations that include inclusive expressions, adding gender-inclusive guidance during its tuning does not harm its overall translation ability. Instead, the model learns to use inclusive language without compromising translation quality. Therefore, targeted gender-inclusive instruction tuning is a practical way to improve the inclusivity of LLMs without sacrificing overall performance.

\paragraph{Key findings} Gender-sensitive translation depends on translation direction and training data size. Translating gender-inclusive Polish into English is relatively straightforward for LLMs, with Polish-specific Bielik-11B performing well even when fine-tuned on the small human-written IPIS test set. In contrast, generating gender-inclusive Polish from English is more challenging and requires larger, more diverse training data. Instruction tuning on the mixed-generated IPIS dataset significantly improves performance, particularly in EN$\rightarrow$PL. Bielik-11B consistently outperforms the SOTA NLLB model, showing the effectiveness of language-specific adaptation. Our results also show that gender inclusivity is not captured by standard MT metrics and can be improved through instruction tuning without reducing overall translation quality.

\subsection{General capabilities of IPIS-tuned LLM}
To evaluate the effect of IPIS-tuning on general knowledge capabilities, we use \textsl{LLMzSzŁ}~\cite{jassem2025llmzsz}, a comprehensive and publicly available Polish benchmark comprising 18.8k closed-ended questions from 154 domains. We compare the \textsl{default} Bielik-11B with its IPIS-tuned variants on proofreading and translation tasks (Table \ref{tab:general_capabilities}).

\begin{table}[!ht]
\centering
{\small
\renewcommand\tabcolsep{14pt}
\begin{tabular}{l||r|r}
\toprule
\textbf{LLM} & \textbf{Accuracy} & \textbf{Delta} \\
\midrule
\textsl{default}               & 56.56 & \multicolumn{1}{c}{$-$} \\
\textsl{tuned (proofreading)}  & 55.74 & -0.82 \\
\textsl{tuned (translation)}   & 55.87 & -0.69 \\
\bottomrule
\end{tabular}}
\caption{Performance of IPIS-tuned Bielik-11B variants vs.\ the default model on \textsl{LLMzSzŁ}.}
\label{tab:general_capabilities}
\end{table}

The default model achieves an accuracy of $56.56\%$, while IPIS-tuned variants show only marginal drops: $55.74\%$ for the proofreading model ($-0.82\%$) and $55.87\%$ for the translation model ($-0.69\%$). Such a minor degradation is expected when fine-tuning for specialised tasks and indicates that IPIS datasets can be readily incorporated into broader instruction corpus without meaningfully affecting downstream performance, while improving model gender inclusivity.

\section{Related work}
\label{sec:sota}
\subsection{Gender bias in language models}
Language models often reproduce gender biases present in their training data \cite{stanczak2021surveygenderbiasnatural,jumelet-etal-2019-analysing,doyen2025manlanguagemodelsevaluating}, resulting in stereotypical, discriminatory or prejudicial outputs that affect applications such as hiring systems \cite{ai5010019}, recommendation systems \cite{cb829bbb6d504bfa9e3d210d506b54b1}, and content generation \cite{chu2024fairnesslargelanguagemodels}. To address this issue, numerous debiasing methods have been proposed \cite[e.g.][]{sun-etal-2019-mitigating, choubey-etal-2021-gfst, saunders-etal-2022-first, borah-mihalcea-2024-towards, NEMANI2024100047}.
Unlike conventional debiasing approaches, which mitigate bias as a post-processing step, we integrate gender inclusivity directly into LLMs through instruction tuning, enabling them to generate gender-inclusive text without post-processing.

Related work has explored gender-fair rewriting and translation. For rewriting, \citet{amrhein-etal-2023-exploiting} proposed a German model trained on automatically constructed data, while \citet{veloso-etal-2023-rewriting} and \citet{piergentili-etal-2025-gender} developed gender-neutral rewriting systems for Portuguese and Italian. In MT, previous work has focused on evaluating gender-fair translation in German \cite{lardelli-etal-2024-building,pranav-etal-2025-glitter} and gender-neutral translation in Italian \cite{savoldi-etal-2021-gender,savoldi-etal-2025-mind,piergentili-etal-2025-llm}. \citet{nunziatini-diego-2024-implementing} used GPT-4 to post-edit DeepL translations into gender-inclusive text. Our work adapts open-source LLMs through instruction tuning to generate gender-inclusive Polish (possibly with gender-neutral expressions), rather than fully gender-neutral language. Our IPIS dataset covers two tasks (proofreading and translation), spans multiple text domains, and is fully expert-annotated.
Research on gender bias in Polish NLP remains scarce. One notable exception is \citet{martinkova-etal-2023-measuring}, who measure gender bias in Polish, Czech and Slovak LMs. Our work complements this line of research by introducing resources and IPIS-tuned models for gender-fair Polish.

\subsection{Instruction tuning}

Instruction tuning is a fine-tuning technique that enhances LLMs' ability to follow instructions and to solve specific tasks
\cite{zhang2024instructiontuninglargelanguage,ustun-etal-2024-aya,ouyang2022traininglanguagemodelsfollow,mishra-etal-2022-cross}. LLMs can be instruction-tuned for NLP and NLG tasks, such as classification \cite{rosenbaum-etal-2022-linguist}, information extraction \cite{Sun_Zhang_Li_Lou_2024}, and writing assistance \cite{chakrabarty-etal-2022-help,zhang2023multitaskinstructiontuningllama,raheja-etal-2023-coedit,raheja-etal-2024-medit}, but also for code generation and solving arithmetic problems \cite{NEURIPS2024_3d5aa9a7}. Moreover, training on diverse instruction datasets improves models' general instruction-following abilities, including on previously unseen tasks \cite{wei2022finetunedlanguagemodelszeroshot}.

Our work is most closely related to instruction-tuned writing assistants, e.g. \textsc{CoEdIT} \cite{raheja-etal-2023-coedit}, \textsc{mEdIT} \cite{raheja-etal-2024-medit}, and instruction-tuned LLaMA models for text editing \cite{zhang2023multitaskinstructiontuningllama}. Unlike these approaches, which focus on simplification, paraphrasing, and style transfer tasks, we use instruction tuning to adapt open-source LLMs for gender-inclusive proofreading and gender-sensitive translation in Polish.

\section{Conclusions}
\label{sec:econclusions}
This study presents a pioneering attempt to integrate gender inclusivity into Polish LLMs through instruction tuning by promoting explicit, context-appropriate gender marking rather than full gender neutralisation. A key challenge of instruction tuning lies in developing high-quality instructional data suitable for LLM training. To address this, we introduced the IPIS dataset, a~human-crafted collection of gender-inclusive proofreading instructions in Polish, and gender-sensitive Polish--English translation instructions. By focusing on a~less commonly studied language -- Polish -- and a novel research area -- gender inclusivity, IPIS enriches existing open-source instruction tuning resources and expands the scope of LLM adaptation.

Our experimental setup involved tuning open-source multilingual LLMs (Llama, Mistral, and Mistral-Nemo) and Polish-specific models (Bielik and PLLuM). Importantly, since we used only open-source LLMs, all experiments can be replicated, ensuring transparency and reproducibility.

The results confirmed that off-the-shelf LLMs inherently lack gender sensitivity because they are trained on non-inclusive textual corpora. The tasks examined in this study -- gender-inclusive proofreading and gender-sensitive translation -- proved highly complex and context-dependent. They require LLMs to reason about referential gender, linguistic structure, and world knowledge to distinguish personal from non-personal nouns, identify generic masculine forms that require modifications, as opposed to genuine masculine forms that should remain unchanged, and generate grammatically and socially appropriate inclusive forms.

Our experiments showed that LLMs can perform the investigated tasks effectively, if they are tuned on a high-quality and sufficiently large set of instructions. In particular, 23k IPIS-proofreading instructions were sufficient to tune LLMs to generate gender-inclusive Polish texts. Limited training data in the preliminary translation experiment, in turn, led to overall model degradation. Bielik-11B fine-tuned on the mixed-generated IPIS-translation train set (23k) outperformed both its preliminary variant and the strong multilingual NLLB model.

LLM-as-a-judge and human evaluation further showed that gender inclusivity is only weakly correlated with \textsc{bleu} and chrF scores. It is an independent dimension of translation quality that can be improved through targeted instruction tuning without reducing overall translation performance. 

Last but not least, evaluation on \textit{LLMzSzŁ} confirmed that IPIS tuning improved gender inclusivity while preserving the LLMs' general capabilities.

Our approach embeds gender inclusivity into LLMs, offering a systematic way to mitigate gender bias in Polish language generation. By advancing gender-inclusive NLP and NLG, it contributes to broader efforts to promote gender equality through more inclusive and equitable communication.

\section*{Limitations}
\label{sec:limitations}
In this work, we develop and evaluate instruction-tuned LLMs for gender-inclusive proofreading in Polish and bidirectional translation between gender-inclusive Polish and English. However, our approach has several limitations that should be acknowledged. An obvious limitation is the linguistic scope of our research, as we focus mostly on Polish. While this limits the generalisability of our findings to other languages, it also reduces the risk that the tested LLMs encountered the IPIS dataset during pretraining, minimising potential data contamination. Additionally, Polish is also particularly challenging for LLMs because of its complex grammatical gender system and strong masculine bias, making it a particularly demanding language for gender-inclusive NLP. Despite this language-specific focus, our work contributes to broader efforts to improve linguistic diversity in the largely English- and Chinese-dominated NLP world.

Our research focuses on binary gender inclusion. Although Polish generic masculine can refer to individuals of any gender, including non-binary individuals, we target only feminine forms, masculine-feminine dyads, and neutral expressions. We do not model non-binary forms, as they have not yet become standardised and remain socially emergent, unlike well-established feminine forms. Following a descriptive rather than prescriptive (normative) approach to language use, we do not introduce non-binary forms into our dataset. We acknowledge that this scope does not ensure full inclusivity, but consider the systematic inclusion of feminine forms linguistically grounded and socially relevant starting point. It is also worth noting that, from a broad repertoire of abbreviated gender-inclusive notations, we adopt one proposed by \citet{wro:inni:2025}, e.g. \textit{urzędni*k/czka/cze} [an official], which encodes masculine (\textit{urzędnik}), feminine (\textit{urzędniczka}), and a candidate non-binary form (\textit{urzędnicze}). In our study, however, we use only its masculine--feminine version, without non-binary forms. This notation is nevertheless extensible and non-binary forms can be incorporated when they become sufficiently established in Polish.

We use \textsc{bleu} and chrF as reference-based indicators of translation quality; however, both metrics have important limitations when evaluating gender-inclusive language. They measure surface-level similarity to reference translations and cannot reliably distinguish between gender-biased and gender-inclusive formulations. In particular, they may penalise valid gender-inclusive or neutral reformulations simply because these differ from the reference, even when they are linguistically correct and preserve the intended meaning. This issue is especially relevant for English$\rightarrow$Polish translation, as English often leaves referential gender unspecified, whereas Polish frequently requires gender to be expressed through grammatical forms. Consequently, multiple valid translations may differ in how they resolve gender ambiguity, including whether they use gendered dyads, feminine forms, or gender-neutral expressions. A single reference translation cannot capture this range of acceptable solutions. Therefore, \textsc{bleu} and chrF should be interpreted as coarse indicators of surface-level correspondence rather than direct measures of gender inclusivity or bias mitigation. To address these limitations, we complement automatic MT metric with an LLM-as-a-judge evaluation and a manual review of 114 translations.

Another limitation relates to the computational resources required for our experiments. We tune multiple LLMs with billions of parameters, and although we employ LoRA to optimise efficiency, the overall computational cost remains substantial. This not only poses a financial barrier to replicating the results due to resource constraints but also has environmental implications. To mitigate this, we publicly release our best models, facilitating further research while reducing redundant costs.

\section*{Acknowledgments}
We gratefully acknowledge Polish high-performance computing infrastructure PLGrid (HPC Center: ACK Cyfronet AGH) for providing computer facilities and support within computational grant no. PLG/2025/018815.\\
The gender-inclusive Polish corpus underlying IPIS was developed as part of the PLLUM project, founded by the Polish Ministry of Digital Affairs (grant no. 1/WI/DBiI/2023).

\bibliography{custom}

\newpage
\appendix

\section{Challenging translation examples}
\label{sec:appendixC2}
Gender-sensitive translation is not a trivial task. To the best of our knowledge, there are no dedicated English$\leftrightarrow$gender-inclusive Polish parallel corpora. As a result, LLMs are usually trained on conventional parallel texts and rarely learn how gender is encoded differently in the two languages.

\subsection{Polish-to-English translation} 
In Polish, gender-inclusive language is often expressed through noun, pronoun, adjective, and verb dyads. These forms are ge\-ne\-rally translated into gender-neutral English expressions, which can lead to significant differences in token count between Polish and English texts, cf. \ref{ex:1} and \ref{ex:1a}. Moreover, because explicitly marked gender-inclusive forms may sound unnatural in English, it remains unclear whether such distinctions should be preserved in translation, see \ref{ex:1b} and \ref{ex:3}, or replaced with more natural gender-neutral alternatives, see \ref{ex:1a} and \ref{ex:4}.

\ex. \label{ex:1}\textit{Szczyt Women In Tech zgromadził \textcolor{Green}{liczne uczestniczki} i \textcolor{NavyBlue}{licznych uczestników}, \textcolor{Green}{które}/\textcolor{NavyBlue}{którzy} \textcolor{Red}{uczestniczyły}/\textcolor{Orange}{uczestniczyli} w różnorodnych prelekcjach}.\\
    {\fontsize{10}{10}\selectfont $[$gloss.$]$ Woman In Tech Summit attracted \textcolor{Green}{numerous$_{\text{fem}}$ attendees$_{\text{fem}}$} and \textcolor{NavyBlue}{numerous$_{\text{masc}}$ attendees$_{\text{masc}}$}\footnote{Both women and men participate in Women In Tech Summits.} \textcolor{Green}{who$_{\text{fem}}$}/\textcolor{NavyBlue}{who$_{\text{masc}}$} \textcolor{Red}{participate$_{\text{fem}}$}/\textcolor{Orange}{participate$_{\text{masc}}$} in a variety of lectures}
    \a. \label{ex:1a}Women In Tech Summit attracted numerous attendees who participated in a variety of lectures.
    \b. \label{ex:1b}$?$Women In Tech Summit attracted numerous female attendees and numerous male attendees who participated in a variety of lectures.
    \c. *Women In Tech Summit attracted numerous attendees and numerous attendees who participated in a variety of lectures.

\ex. \textit{W piątek policja aresztowała \textcolor{NavyBlue}{pracownika} i \textcolor{Green}{pracownicę} banku.}\\
    {\fontsize{10}{10}\selectfont $[$gloss.$]$ On Friday the police arrested \textcolor{NavyBlue}{employee$_{\text{masc}}$} and \textcolor{Green}{employee$_{\text{fem}}$} of bank}
    \a.\label{ex:4} On Friday, the police arrested two bank employees.
    \b.\label{ex:3} ?On Friday, the police arrested a male and a female bank employee.

The aim of Polish$\rightarrow$English translation experiments is to evaluate whether LLMs can generate high-quality English translations from gender-inclusive Polish source texts. Although the Polish input contains gendered dyads, the expected outputs are natural gender-neutral English translations without explicit gender marking, redundant repetitions, or other unnecessary linguistic phenomena.

\subsection{English-to-Polish translation}

Depending on the context, English gender-neutral expressions should be translated into Polish either as gendered dyads, see \ref{ex:0a} and \ref{ex:5a}, or as single gendered forms, see \ref{ex:6a} and the feminine translation of `Rector' in \ref{ex:0a}. Translations with generic masculine forms are not acceptable, see \ref{ex:0b}, \ref{ex:5c}, and \ref{ex:6c}, nor are translations that contradict world knowledge, see \ref{ex:5b} and \ref{ex:6b}.

 \ex. \label{ex:0} \textit{The most outstanding students were awarded by the Rector of AMU.}
    \a. \label{ex:0a}\textcolor{NavyBlue}{Najwybitniejsi studenci} i \textcolor{Green}{naj\-wy\-bit\-niejsze studentki} \textcolor{Orange}{zostali nagrodzeni}/\textcolor{Red}{zostały nagrodzone} przez \textcolor{Green}{Rektorkę} UAM.\\
    {\fontsize{10}{10}\selectfont $[$gloss.$]$ \textcolor{NavyBlue}{The most outstanding$_{\text{masc}}$ students$_{\text{masc}}$} and \textcolor{Green}{the most outstanding$_{\text{fem}}$ students$_{\text{fem}}$} \textcolor{Orange}{were$_{\text{masc}}$ awarded$_{\text{masc}}$}/\textcolor{Red}{were$_{\text{fem}}$ awarded$_{\text{fem}}$} by \textcolor{Green}{the Rector$_{\text{fem}}$} of AMU}
    \b. \label{ex:0b}*\textcolor{NavyBlue}{Najwybitniejsi studenci} \textcolor{Orange}{zostali nagrodzeni} przez \textcolor{NavyBlue}{Rektora} UAM.\\
    {\fontsize{10}{10}\selectfont $[$gloss.$]$ \textcolor{NavyBlue}{The most outstanding$_{\text{masc}}$ students$_{\text{masc}}$}\footnote{Both women and men can study at AMU.} \textcolor{Orange}{were$_{\text{masc}}$ awarded$_{\text{masc}}$} by \textcolor{NavyBlue}{the Rector$_{\text{masc}}$}\footnote{The Rector of AMU is Her Magnificence prof. dr hab. Bogumiła Kaniewska.} of AMU}

\ex. \label{ex:5}\textit{Patients rated Eye Clinic positively}
    \a. \label{ex:5a}\textcolor{NavyBlue}{Pacjenci} i \textcolor{Green}{pacjentki} pozytywnie \textcolor{Orange}{ocenili}/\textcolor{Red}{oceniły} Eye Clinic.\\
    {\fontsize{10}{10}\selectfont $[$gloss.$]$ \textcolor{NavyBlue}{Patients$_{\text{masc}}$} and \textcolor{Green}{patients$_{\text{fem}}$} positively \textcolor{Orange}{rated$_{\text{masc}}$}/\textcolor{Red}{rated$_{\text{fem}}$} Eye Clinic}\\
    \b. \label{ex:5c}*\textcolor{NavyBlue}{Pacjenci} pozytywnie \textcolor{Orange}{ocenili} Eye Clinic.\\ 
    {\fontsize{10}{10}\selectfont $[$gloss.$]$ \textcolor{NavyBlue}{Patients$_{\text{masc}}$} positively \textcolor{Orange}{rated$_{\text{masc}}$} Eye Clinic}\\
    \c. \label{ex:5b}*\textcolor{Green}{Pacjentki} pozytywnie \textcolor{Red}{oceniły} Eye Clinic.\\
    {\fontsize{10}{10}\selectfont $[$gloss.$]$ \textcolor{Green}{Patients$_{\text{fem}}$}\footnote{Both women and men may receive treatment in Eye Clinic, and it is likely that individuals of both genders have provided ratings for the clinic.} positively \textcolor{Red}{rated$_{\text{fem}}$} Eye Clinic}

\ex. \label{ex:6} \textit{Patients rated Medifem positively.}
    \a. \label{ex:6a}\textcolor{Green}{Pacjentki} pozytywnie \textcolor{Red}{oceniły} Medifem.\\ 
    {\fontsize{10}{10}\selectfont $[$gloss.$]$ \textcolor{Green}{Patients$_{\text{fem}}$} positively \textcolor{Red}{rated$_{\text{fem}}$} Medifem}
    \b. \label{ex:6c}*\textcolor{NavyBlue}{Pacjenci} pozytywnie \textcolor{Orange}{ocenili} Medifem.\\
    {\fontsize{10}{10}\selectfont $[$gloss.$]$ \textcolor{NavyBlue}{Patients$_{\text{masc}}$} positively \textcolor{Orange}{rated$_{\text{masc}}$} Medifem}\\
    \c. \label{ex:6b}*\textcolor{NavyBlue}{Pacjenci} i \textcolor{Green}{pacjentki} pozytywnie \textcolor{Orange}{ocenili}/\textcolor{Red}{oceniły} Medifem.\\
    {\fontsize{10}{10}\selectfont $[$gloss.$]$ \textcolor{NavyBlue}{Patients$_{\text{masc}}$}\footnote{Medifem is a women's clinic, so men cannot be its patients.} and \textcolor{Green}{patients$_{\text{fem}}$} positively \textcolor{Orange}{rated$_{\text{masc}}$}/\textcolor{Red}{rated$_{\text{fem}}$} Medifem}

The English$\rightarrow$Polish translation experiments examine whether LLMs can translate English texts with minimal gender cues into gender-inclusive Polish by making contextually appropriate and inclusive choices in the target language.

\section{Details on the IPIS dataset}
\label{sec:appendix0}
The IPIS dataset is publicly available under the CC BY-SA 4.0 licence at \url{https://huggingface.co/datasets/ipipan/IPIS}.

\subsection{Text sources in IPIS}
\label{sec:apx:ipis_content}

\paragraph{Legal and administrative texts} (LEGAL)
    \begin{itemize}
        \item European legal acts (Polish and English): \href{https://www.europarl.europa.eu/charter/pdf/text_en.pdf}{\textit{Charter of Fundamental Rights of the European Union}}, \href{https://eur-lex.europa.eu/resource.html?uri=cellar:2bf140bf-a3f8-4ab2-b506-fd71826e6da6.0023.02/DOC_1&format=PDF}{\textit{The Treaty on European Union}} (Consolidated Version), and \href{https://eur-lex.europa.eu/LexUriServ/LexUriServ.do?uri=CELEX:12012E/TXT:en:PDF}{\textit{The Treaty on the Functioning of the European Union}} (Consolidated Version).
        \item Polish legal acts (Polish): \textit{\href{https://isap.sejm.gov.pl/isap.nsf/download.xsp/WDU19970780483/U/D19970483Lj.pdf}{Konstytucja Rzeczypospolitej Polskiej}} (Eng. Constitution of Poland).
        \item Institutional documents (Polish): internal administrative documents of ICS PAS.
    \end{itemize}

\paragraph{National Corpus of Polish}(\href{https://huggingface.co/datasets/clarin-pl/nlprepl}{NKJP1M}): the 1-million, balanced sub-corpus of NKJP \cite{przepiorkowski2012nkjp} using the NLPre-based split \cite{wiacek-etal-2024-nlpre}.

\paragraph{Polish parliamentary transcripts} (\href{https://www.clarin.si/repository/xmlui/handle/11356/1860}{TRANSCRIPT}): a subset of \href{https://www.clarin.eu/parlamint}{ParlaMint} \cite{erjavec2023parlamint}, i.e. a corpus of transcribed parliamentary debates in Polish.
 
\subsection{Human annotation of IPIS}
\label{sec:apx:ipis_details}

This section presents the human annotation process used to create the IPIS underlying corpus.

\paragraph{Annotators.} The annotation was performed by trained Polish linguists (native Polish speakers) employed full-time in a project. The team included two postdoctoral researchers and four PhD students.

\paragraph{Annotation procedure.}
Each text was independently transformed by two annotators. A third annotator reviewed both versions and could accept them, select one, or revise one or both proposals.

\paragraph{Gender-inclusive strategies.}
Annotators used two main inclusive strategies: feminisation (introducing feminine forms) and neutralisation (using impersonal forms or collective nouns). No single strategy was enforced; annotators selected the most appropriate strategy depending on the context (see Section \ref{sec:apx:strategies_illustrations} for examples that illustrate gender-inclusive strategies).

\paragraph{Mixing strategies.}
Annotators were allowed to combine different strategies within the same text. This reflects natural language use, where such variation is common. As a result, we did not require consistent use of a single strategy, nor did we ask to create multiple versions of a text for each strategy. Instead, the annotators produced a single fluent version adapted to the genre.

\paragraph{Annotation guidelines.}
Annotators followed se\-ve\-ral general guidelines:
\begin{itemize}
    \item Different inclusive strategies and notation variants could be combined within a single text if appropriate for the genre (e.g. gender-star notation should be avoided in conversation transcripts).
    \item Gender-star notation was the only allowed abbreviated form.
    \item Annotators were encouraged to follow the principle of linguistic economy, e.g. using a full gendered dyad at first mention and an abbreviated form later.
    \item Grammatical agreement between nouns, verbs, and adjectives had to be consistently adjusted; gender-star expressions are preferred in such cases.
\end{itemize}

\paragraph{Task difficulty.}
Annotators did not create new lexical items, but used existing, well-established forms. As native speakers, they generally did not find the task of introducing gender-inclusive or gender-neutral nouns, or adjusting grammatical agreement, particularly difficult.

\renewcommand*{\arraystretch}{1.3}
\renewcommand\tabcolsep{6.5pt}
\begin{table*}[!ht]
\small
\centering
\begin{tabular}{c||lr|rr|rr|r}
    \toprule
\multirow{2}{*}{\textbf{Domain}} & \multirow{2}{*}{\textbf{Level}} & \multirow{2}{*}{\textbf{Total}} & \multicolumn{2}{c|}{\textbf{Annotation1}} & \multicolumn{2}{c|}{\textbf{Annotation2}}  & 
\multirow{2}{*}{\parbox[c]{1.5cm}{\centering\textbf{Both}}}\\
&&&\parbox[c]{1.5cm}{\centering\textbf{Count}} & \textbf{Agr(\%)} & \parbox[c]{1.5cm}{\centering\textbf{Count}} & \textbf{Agr(\%)} &\\
\midrule
\multirow{2}{*}{\textbf{LEGAL}} & txt &  4755 & 1,220 (25.7\%) & 99.3 & 1,221 (25.7\%) & 99.3 & 1,212 (25.5\%) \\
& tok & 97,361 &  2,925 (3.0\%) & 98.9  & 2,913 (3.0\%) & 99.3 & 2,894 (3.0\%)\\
\midrule
\multirow{2}{*}{\textbf{NKJP1M}} & txt & 18,675 & 8,151 (43.6\%) & 90.8 & 8,031 (43.0\%) & 92.2 & 7,402 (39.6\%) \\
& tok & 1,212,528 & 22,434 (1.8\%) & 83.8 & 20,996 (1.7\%) & 89.6 & 18,809 (1.5\%)\\
\midrule
\multirow{2}{*}{\textbf{TRANSCRIPT}} & txt & 6,333 & 4,097 (64.7\%) & 97.3 & 4,101 (64.8\%) & 97.2 & 3,985 (62.9\%) \\
& tok & 811,764 & 16,216 (2.0\%) & 88.5 &  16,330 (2.0\%) & 87.8 & 14,346 (1.8\%)\\
\midrule
\midrule
\multirow{2}{*}{\textbf{ALL TEXTS}} & txt & 29,763 & 13,468 (45.2\%) & 93.5 & 13,353 (44.9\%) & 94.3 & 12,599 (42.3\%)\\
& tok & 2,121,653 & 41,575 (2.0\%) & 86.7 & 40,239 (1.9\%) & 89.6 & 36,049 (1.7\%)\\
\bottomrule
\end{tabular}
\caption{Annotation statistics for the IPIS underlying corpus. \textbf{txt} and \textbf{tok} denote the number of modified texts and tokens, respectively. \textbf{Annotation1} and \textbf{Annotation2} denote modifications introduced by Annotator~1 and Annotator~2, and \textbf{Both} denotes modifications identified by both annotators. \textbf{Agr} denotes the agreement ratio. \textbf{LEGAL} refers to legal and administrative texts, \textbf{NKJP1M} to texts from the balanced, multi-domain National Corpus of Polish, and \textbf{TRANSCRIPT} to Polish parliamentary transcripts.}
\label{tab:annotation_statistics}
\end{table*}

\paragraph{Limitations.}
The annotation process has several limitations:
\begin{itemize}
    \item Because multiple formulations may be equally valid, token-level inter-annotator agreement cannot be reliably measured. Semantically equivalent expressions may differ in form, whereas standard agreement metrics rely on surface similarity.
    \item The annotation process is not fully deterministic, since various inclusive forms and notations are acceptable. However, identifying which elements require gender-related changes remains consistent.
    \item We did not conduct user studies on perceived inclusiveness. Nevertheless, some forms (e.g. feminine forms and gendered dyads) are generally more accepted in Polish than strongly neutralised expressions such as \textit{osoba studencka} [a student person].
\end{itemize}

\subsection{Examples illustrating gender-inclusive strategies}
\label{sec:apx:strategies_illustrations}

\paragraph{Feminisation}
Introducing feminine forms either as gendered dyads (see \ref{ex:fem:1}--\ref{ex:fem:3}) or as replacements for generic masculine forms (see \ref{ex:fem:4}).

\begin{enumerate}
    \item \label{ex:fem:1}Masculine and feminine nouns connected with a coordinating conjunction, e.g. \textit{pacjenci i pacjentki} [patients$_{\text{masc}}$ and patients$_{\text{fem}}$],
    \item \label{ex:fem:2}Masculine and feminine nouns connected with a slash, e.g. \textit{student/studentka} [student$_{\text{masc}}$/student$_{\text{fem}}$],
    \item \label{ex:fem:3}Abbreviated gender-inclusive form with an asterisk (gender star), e.g. \textit{pracowni*ków/czek} [employees$_{\text{masc/fem}}$],
    \item \label{ex:fem:4} A feminine form denoting a function, profession, or position held by a woman, e.g. \textit{dyrektorka Kwiatkowska} [director$_{\text{fem}}$ Kwiatkowska$_{\text{fem}}$] instead of \textit{dyrektor Kwiatkowska} [Director$_{\text{masc}}$ Kwiatkowska$_{\text{fem}}$].
\end{enumerate}

\paragraph{Neutralisation}
Using impersonal expressions (see \ref{ex:neut:1}) or collective nouns (see \ref{ex:neut:2}).

\begin{enumerate}
    \item \label{ex:neut:1}The noun \textit{osoba} [person$_{\text{fem}}$] and its modifiers, e.g. \textit{osoby uczestniczące w spotkaniu} [persons participating in the meeting] instead of \textit{uczestnicy spotkania} [meeting participants$_{\text{masc}}$],
    \item \label{ex:neut:2}A gender-neutral collective noun, e.g. \textit{personel medyczny} [medical staff] instead of \textit{lekarze i pielęgniarki} [doctors$_{\text{masc}}$ and nurses$_{\text{fem}}$]; \textit{ludzie sportu} [sportspeople] instead of \textit{sportowcy} [sportsmen].
\end{enumerate}

\subsection{IPIS quality and statistical summary}
\label{sec:apx:ipis_statistical_summary}

\paragraph{Annotation statistics}

Table \ref{tab:annotation_statistics} presents the annotation statistics. Annotators introduced gender-inclusive modifications most frequently in parliamentary transcripts, where nearly $65\%$ of the text passages were revised. In contrast, legal and administrative texts (LEGAL) required the fewest modifications.

Although the LEGAL subset contains the fewest modified passages, it has the highest proportion of modified tokens (about $3\%$ of all tokens) and the highest agreement ratio (i.e. the proportion of each annotator's modifications that were also identified by the other annotator), reaching approx. $99\%$. Overall, agreement between the two annotators is consistently high across all domains. The lowest agreement ($83.3\%$) is observed for Annotation~1 in the NKJP1M subset, which contains linguistically diverse texts.

\paragraph{Dataset statistics}

Table \ref{tab:annotation_strategies} reports the distribution of gender-inclusive strategies in the IPIS dataset. Since each text passage may have one or two valid annotated versions, we report the statistics for both annotation sets. For each generic masculine expression, the annotators could apply either feminisation or neutralisation.

\renewcommand*{\arraystretch}{1.4}
\renewcommand\tabcolsep{2.5pt}
\begin{table}[!ht]
\centering
\small
\begin{tabular}{l|lrrr}
\toprule
\textbf{Strategy}  & \multicolumn{1}{c}{} & \textbf{Test} & \textbf{Dev} & \textbf{Train} \\
\midrule
\multirow{2}{*}{\textbf{Coord}} & a1 & 3,946 (51.0\%) & 1,546 (27.3\%) & 9,596 (38.4\%)\\
& a2 & 3,300 (42.2\%) & 1,747 (31.4\%) & 9,996 (41.7\%) \\
 \midrule
\multirow{2}{*}{\textbf{Slash}} & a1 & 915 (11.8\%)& 625 (11.1\%)  &  4,193 (16.8\%)\\
& a2 & 1,628 (20.9\%) & 611 (11.0\%) & 4,519 (18.8\%) \\
\midrule
\multirow{2}{*}{\textbf{Star}} & a1 & 1,300 (16.8\%) & 1,546 (27.4\%) & 6,845 (27.4\%) \\
& a2 & 1,314 (16.9\%) & 1,277 (23.0\%) &  5,357 (22.3\%)\\
\midrule
\multirow{2}{*}{\textbf{N/F}} & a1 & 1,574 (20.3\%) & 1,934 (34.2\%) & 4,333 (17.3\%)\\
& a2 & 1,551 (19.9\%)& 1,927 (34.6\%) & 4,112 (17.1\%)\\
\bottomrule
\end{tabular}
\caption{Distribution of gender-inclusive strategies across the IPIS splits (counted by modified spans). Feminisation strategies include coordination (\textbf{Coord}), slash notation (\textbf{Slash}), and gender star notation (\textbf{Star}). \textbf{F/N} denotes feminine or neutral expressions. a1 and a2 denote annotations~1 and~2, respectively. }
\label{tab:annotation_strategies}
\end{table}

Feminisation is realised either by introducing gendered dyads (coordination of masculine and feminine forms, slash notation, or gender-star notation) or by replacing a masculine form with a feminine one. However, surface morphological features cannot distinguish feminine forms referring to women (e.g. \textit{nauczycielka} [teacher$_{\text{fem}}$]) from epicene nouns used as gender-neutral expressions (e.g. \textit{osoba} [person$_{\text{fem}}$]), as both have feminine grammatical gender and share the same morphological features. Therefore, these expressions are grouped into a single N/F category.

All gender-inclusive strategies are well represented in IPIS, although coordination is by far the most frequent. Annotations~1 contain slightly more instances of gender-star notation, whereas Annotations~2 use slash notation somewhat more often. The two annotations show almost identical proportions of N/F expressions, indicating a high level of consistency in strategy selection.

\subsection{IPIS instances}
\label{sec:apx:ipis_instances}
\paragraph{IPIS-proofreading example} (see Figure \ref{fig:apx:ipis-proofreading})

\begin{figure}[!ht]

\begin{minipage}{\textwidth}
{\small
\begin{verbatim}
{'source_resource_id': 'nlprepl-nkjp1m-dev',
 'ipis_id': 'IPIS_proofreading_dev_2143',
 'prompt': 'Przeredaguj tekst w standardowym
    języku polskim, aby nie zawierał treści
    krzywdzących i wykluczających. Tekst 
    źródłowy: ',
 'source': 'W 24-osobowym składzie sędziowskim
    znalazło się 8 Polaków, pianistów i
    pedagogów. W przypadku własnych
    uczniów nie będą mieli prawa głosu.
    Pojawią się też m.in. laureat nagrody
    za najlepsze nagranie Roku Chopinowskiego
    - Emanuel Ax i rosyjska pianistka
    reprezentująca USA - Bella Davidovich.',
 'target': 'W 24-osobowym składzie sędziowskim
    znalazło się 8 Pol*aków/ek, pianist*ów/ek i
    pedago*gów/żek. W przypadku własnych
    ucz*niów/ennic nie będą mi*eli/ały prawa
    głosu.Pojawią się też m.in. laureat nagrody
    za najlepsze nagranie Roku Chopinowskiego
    - Emanuel Ax i rosyjska pianistka
    reprezentująca USA - Bella Davidovich.'}
\end{verbatim}
}
\end{minipage}
\caption{IPIS-proofreading example. [Eng. The 24-member jury includes 8 Polish pianists and educators. They will be excluded from voting for their own students. Among the jurors are also Emanuel Ax, the winner of the award for the best recording of the Chopin Year, and Bella Davidovich -- a Russian-born pianist representing the USA.]}
\label{fig:apx:ipis-proofreading}
\end{figure}

\noindent
Each IPIS-proofreading sample consists of three components:
\begin{enumerate}
    \item \textbf{\texttt{prompt}} -- user prompt, i.e. a specification of the given task (e.g. 'Revise the text in standard Polish so that it does not contain any harmful or exclusionary content. Source text:'),
    \item \textbf{\texttt{source}} -- input text passage, i.e. a text passage requiring a gender-inclusive proofreading,
    \item \textbf{\texttt{target}} -- desired output, i.e. the expected response corresponding to a user instruction and input text passage. This serves as the ground truth for evaluating and optimising LLMs.
\end{enumerate}

\paragraph{IPIS-translation example} (see Figure \ref{fig:apx:ipis-translation})
\begin{figure}[!ht]
\begin{minipage}{\textwidth}
{\small
\begin{verbatim}
{"source_resource_id":"EU_Karta_Praw_Podstawowych",
 "ipis_id":"IPIS-EN_translation_pl2en_dev_39",
 "prompt":"Generate an English translation
    of the text:",
 "source":"IV
    SOLIDARNOŚĆ
    Artykuł 27
    Prawo pracowników i pracownic do informacji
    i konsultacji w ramach przedsiębiorstwa
    Pracownikom/Pracownicom i ich 
    przedstawicielom/przedstawicielkom należy
    zagwarantować, na właściwych poziomach,
    informację i konsultację we właściwym czasie,
    w przypadkach i na warunkach przewidzianych
    w prawie Unii oraz ustawodawstwach
    i praktykach krajowych.",
 "target":"TITLE IV
    SOLIDARITY
    Article 27
    Workers' right to information 
    and consultation within the undertaking
    Workers or their 
    representatives must,
    at the appropriate levels, be guaranteed
    information and consultation in good time
    in the cases and under the conditions provided
    for by Union law and national laws
    and practices.",
 "prompt_language":"EN",
 "source_language":"PL",
 "target_language":"EN",
 "human":true}
\end{verbatim}
}
\end{minipage}
\caption{IPIS-translation example: English$\rightarrow$gender-inclusive Polish translation direction and an user prompt in English.}
\label{fig:apx:ipis-translation}
\end{figure}

\noindent
Each IPIS-translation sample consists of three main components and four translation-specific ones:
\begin{enumerate}
    \item \textbf{\texttt{prompt}} -- user prompt, i.e. a specification of the given task (e.g. 'Generate an English translation of the text:'),
    \item \textbf{\texttt{source}} -- input text passage, i.e. a text passage requiring a gender-sensitive translation,
    \item \textbf{\texttt{target}} -- desired output, i.e. the expected response corresponding to a user instruction and input text passage. This serves as the ground truth for evaluating and optimising LLMs,
    \item \textbf{\texttt{prompt\_language}} -- the language of an user prompt (either EN or PL),
    \item \textbf{\texttt{source\_language}} -- the language of a passage to translate, either gender-inclusive Polish (PL) or standard English (EN),
    \item \textbf{\texttt{target\_language}} -- the language of a reference translation, either standard English (EN) or gender-inclusive Polish (PL),
    \item \textbf{\texttt{human}} -- the component of the full mixed-annotated set indicating whether the English text passage was written by a human (\texttt{true}) or not (\texttt{false}) (i.e. it was translated by the Gemma-3-27B model).
\end{enumerate}

\section{System and evaluation prompts}
\label{sec:appendixA}
\begin{enumerate}
\item System prompt for gender-inclusive proofreading -- Figure \ref{fig:system_prompt_proofreading}
\item System prompt for gender-sensitive translation -- Figure \ref{fig:system_prompt_translation}
\item Evaluation prompt for LLM-as-a-judge evaluation -- Figure \ref{fig:llm-as-a-judge-prompt}
\end{enumerate}
\begin{figure*}
\fbox{
\fontsize{8}{9}\selectfont
\begin{minipage}{0.97\textwidth}
You are an editor, who verifies whether a Polish text is written in gender-inclusive language. If not, you should transform it according to the principles of gender-inclusive language and the text's genre.\\

Follow this algorithm:

1. Identify the genre of the text and the forms of gender-inclusive expressions allowed in it.

2. Identify exclusionary expressions (see section I).

3. Replace exclusionary expressions with gender-inclusive expressions appropriate to the genre (see II).

4. Transform the grammatical forms of dependent expressions to agree with the gender-inclusive expression so that the text remains grammatically correct (see III).\\

I. EXCLUSIONARY EXPRESSIONS IN POLISH

a) An expression in the generic masculine gender naming mixed-gender groups [e.g. 'pacjenci' - both women (pacjentki) and men (pacjenci) can fall ill].\\

b) An expression in the generic masculine gender naming an unspecified person from a mixed-gender group [e.g. 'student' – a woman (studentka) and a man (student) may study].\\

c) A masculine expression naming a function, profession, or position a woman holds [e.g. 'dyrektor Kwiatkowska' – Kwiatkowska should be referred to as dyrektorka].\\

II. INCLUSIVE EXPRESSIONS AND TEXT GENRE\\

a) Masculine and feminine nouns with a coordinating conjunction

- a feminine and a masculine noun connected by a coordinating conjunction

- e.g. 'pacjenci i pacjentki'

- speeches and oral statements, expressions to an addressee in written documents, narrative texts, scientific and press articles\\

b) Masculine and feminine nouns with a slash

- a feminine and a masculine form connected by a slash (/)

- e.g. 'student/studentka'

- forms and documents, legal acts, speeches and oral statements, narrative texts, scientific and press articles\\

c) Abbreviated gender-inclusive form with an inclusive asterisk

- the expression consists of a root (i.e. the shared part of masculine and feminine forms), an inclusive asterisk (*), and gender-specific suffixes connected by a slash in the predefined male/female order or only the female suffix if the male suffix is null

- e.g. 'pracowni*ków/czek'

- forms and documents, legal acts, narrative texts, scientific and press articles\\

d) Osoba-form

- the noun 'osoba' (Eng. person) and its attribute

- 'osoby uczestniczące w spotkaniu' instead of 'uczestnicy spotkania'

- forms, documents, legal acts\\

e) Neutral form

- a gender-neutral collective noun

- 'personel medyczny' instead of 'lekarze i pielęgniarki'

- forms, documents, legal acts\\

III. GRAMMATICAL AGREEMENTS\\

a) You MUST transform the grammatical gender of the modifier to agree with the gender of the governing gender-inclusive expression [e.g. pracowni*cy/czki naukow*i/e].\\

b) If the gender-inclusive expression is the subject of a sentence, you MUST adjust the grammatical gender of the predicate to agree with the gender of that subject [e.g. rolniczki/rolnicy strajkowały/strajkowali].\\

c) You MUST transform the grammatical gender of pronouns referring to the gender-inclusive expression throughout the text [e.g. Student*ka ma obowiązki. Jego/jej absencja jest nieakceptowana].\\

d) You MUST use collective numerals when referring to mixed-gender groups [e.g. pięcioro kandydat*ów/ek].\\

It is NOT ALLOWED to paraphrase or edit parts of the text that are not exclusionary expressions.\\

It is NOT ALLOWED to add new text that does not result from inclusive transformations.
\end{minipage}}
\caption{System prompt in English for the gender-inclusive proofreading task.}
\label{fig:system_prompt_proofreading}
\end{figure*}

\begin{figure*}
\fbox{
\fontsize{8}{9}\selectfont
\begin{minipage}{0.97\textwidth}
You are a Polish translator.\\

To translate texts written in English into gender-inclusive Polish, follow this algorithm:

1. Specify the genre of the text and the forms of gender-inclusive expressions allowed in it.

2. Translate English text into gender-inclusive Polish.

3. If you identify exclusionary expressions (see section I) in the translation:

    - replace exclusionary expressions with gender-inclusive expressions appropriate to the genre (see II).
    
    - transform the grammatical forms of dependent expressions to agree with the gender-inclusive expression so that the text remains grammatically correct (see III).\\

To translate gender-inclusive Polish texts into English, follow this algorithm:

1. Specify the genre of the text and the forms of Polish gender-inclusive expressions allowed in it (see II).

2. Identify gender-inclusive expressions in the Polish source text.

3. Translate gender-inclusive expressions into their corresponding English expressions [e.g. 'ambitni studenci i ambitne studentki' -> 'ambitious students'].\\

I. EXCLUSIONARY EXPRESSIONS IN POLISH\\

a) An expression in the generic masculine gender naming mixed-gender groups [e.g. 'pacjenci' - both women (pacjentki) and men (pacjenci) can fall ill].\\

b) An expression in the generic masculine gender naming an unspecified person from a mixed-gender group [e.g. 'student' – a woman (studentka) and a man (student) may study].\\

c) A masculine expression naming a function, profession, or position a woman holds [e.g. 'dyrektor Kwiatkowska' – Kwiatkowska should be referred to as dyrektorka].\\

II. INCLUSIVE EXPRESSIONS AND TEXT GENRE\\

a) Masculine and feminine nouns with a coordinating conjunction

- a feminine and a masculine noun connected by a coordinating conjunction

- e.g. 'pacjenci i pacjentki'

- speeches and oral statements, expressions to an addressee in written documents, narrative texts, scientific and press articles\\

b) Masculine and feminine nouns with a slash

- a feminine and a masculine form connected by a slash (/)

- e.g. 'student/studentka'

- forms and documents, legal acts, speeches and oral statements, narrative texts, scientific and press articles\\

c) Abbreviated gender-inclusive form with an inclusive asterisk

- the expression consists of a root (i.e. the shared part of masculine and feminine forms), an inclusive asterisk (*), and gender-specific suffixes connected by a slash in the predefined male/female order or only the female suffix if the male suffix is null

- e.g. 'pracowni*ków/czek'

- forms and documents, legal acts, narrative texts, scientific and press articles\\

d) Osoba-form

- the noun 'osoba' (Eng. person) and its attribute

- 'osoby uczestniczące w spotkaniu' instead of 'uczestnicy spotkania'

- forms, documents, legal acts\\

e) Neutral form

- a gender-neutral collective noun

- 'personel medyczny' instead of 'lekarze i pielęgniarki'

- forms, documents, legal acts\\

III. GRAMMATICAL AGREEMENTS\\

a) You MUST transform the grammatical gender of the modifier to agree with the gender of the governing gender-inclusive expression [e.g. pracowni*cy/czki naukow*i/e].\\

b) If the gender-inclusive expression is the subject of a sentence, you MUST adjust the grammatical gender of the predicate to agree with the gender of that subject [e.g. rolniczki/rolnicy strajkowały/strajkowali].\\

c) You MUST transform the grammatical gender of pronouns referring to the gender-inclusive expression throughout the text [e.g. Student*ka ma obowiązki. Jego/jej absencja jest nieakceptowana].\\

d) You MUST use collective numerals when referring to mixed-gender groups [e.g. pięcioro kandydat*ów/ek].\\

It is NOT ALLOWED to paraphrase or edit parts of the text that are not exclusionary expressions.\\

It is NOT ALLOWED to add new text that does not result from inclusive transformations.
\end{minipage}}
\caption{System prompt in English for the gender-sensitive translation task.}
\label{fig:system_prompt_translation}
\end{figure*}

\begin{figure*}
\fbox{
\fontsize{8}{9}\selectfont
\begin{minipage}{0.97\textwidth}
You are an expert evaluator of translation quality. Your task is to assess a machine-generated Polish translation by comparing it with a reference Polish translation.\\

Evaluate whether the generated translation adheres to standard Polish language conventions, including appropriate terminology, grammar, and stylistic and rhetorical norms. Where applicable, also assess compliance with gender-inclusive language conventions used in the reference -- applying inclusive forms where necessary, but avoiding their unjustified addition.\\

---\\

\#\# Evaluation Procedure\\

\#\#\# Step 1 -- Identify Issues\\

Carefully compare the machine translation with the reference. For each issue found, provide:

- a brief explanation of the problem

- the relevant fragment from both the translation and the reference\\

Use the following issue types:\\

- **Omission** -- missing clauses, sentences, or details

- **Addition** -- information present in the translation but absent from the reference

- **Meaning error** -- distorted or incorrect meaning relative to the reference

- **Terminology** -- incorrect Polish terminology or mistranslations

- **Structure** -- changes in paragraph structure or enumeration layout

- **Style** -- deviations from standard Polish style or awkward phrasing

- **Gender inclusivity** -- missing, incorrect, or inconsistent use of gender-inclusive forms\\

---\\

\#\#\# Step 2 -- Assign Scores\\

Score each criterion from **1** (very poor) to **5** (excellent).\\

**1. Completeness**

Does the translation include all information present in the reference?

- Missing sentences, clauses, or phrases

- Unjustified additions not present in the reference

- Redundant repetitions\\

**2. Clarity and Logical Coherence**

Is the translation of the Polish text clear and logically coherent?

- Written entirely in Polish

- Logically structured at both the sentence and text levels

- Free of incorrect or misleading information\\

**3. Structural Fidelity**

Does the translation preserve the structural organisation of the reference?

- Paragraph and enumeration layout

- References to other texts\\

**4. Linguistic Correctness**

Is the translation grammatically correct and stylistically appropriate?

- Polish orthography, punctuation, and grammar

- Stylistic and phraseological accuracy

- Appropriate lexical choices and phrasing\\

**5. Gender-Inclusive Language**

Does the translation correctly reflect the gender-inclusive conventions of the reference?

- Preservation of inclusive forms (e.g., *członkowie i członkinie*, *członk*owie/inie*)

- Consistency in applying inclusive forms

- Absence of unjustified gender bias or unjustified inclusive forms\\

---\\

\#\#\# Step 3 -- Output\\

Return your evaluation using the following JSON structure only -- no additional commentary:\\

\end{minipage}}
\caption{LLM-as-a-judge prompt for evaluation of English$\rightarrow$Polish translation.}
\label{fig:llm-as-a-judge-prompt}
\end{figure*}

\newpage
\section{List of tested LLMs}
\label{sec:appendix7a}
\paragraph{Tested multilingual LLMs}
\begin{enumerate}
\item \textbf{Llama-8B}\footnote{\fontsize{7}{9}{\url{https://huggingface.co/meta-llama/Llama-3.1-8B-Instruct}}} \cite{grattafiori2024llama3herdmodels}, a~8~billion-parameter model with a~context length of 8k tokens,
\item \textbf{Mistral-7B}\footnote{\fontsize{7}{9}{\url{https://huggingface.co/mistralai/Mistral-7B-Instruct-v0.2}}} \cite{jiang2023mistral7b}, a 7 billion-parameter model,
\item \textbf{Mistral-Nemo}\footnote{\fontsize{7}{9}{\url{https://huggingface.co/mistralai/Mistral-Nemo-Instruct-2407}}} \cite{mistral_nemo}, a 12 billion-parameter model with a a~context window of 128k tokens. 
\end{enumerate}

\paragraph{Tested Polish-specific LLMs}
Polish-specific LLMs were optimised to enhance performance on tasks that require a deeper understanding of Polish.
\begin{enumerate}
    \item \textbf{Bielik-7B}\footnote{\fontsize{7}{9}{\url{https://huggingface.co/speakleash/Bielik-7B-Instruct-v0.1}}} \cite{ociepa2024bielik7bv01polish}, a 7~billion-parameter model derived from Mistral-7B-v0.1,
    \item \textbf{Llama-PLLuM-8B}\footnote{\fontsize{7}{9}{\url{https://huggingface.co/CYFRAGOVPL/Llama-PLLuM-8B-chat}}} \cite{pllum2025}, a~11~billion-parameter variant based on Mixtral 8x22B, 
    \item \textbf{Bielik-11B}\footnote{\fontsize{7}{9}{\url{https://huggingface.co/speakleash/Bielik-11B-v2.3-Instruct}}} \cite{Bielik11Bv21i} built upon Llama3.1-8B,
    \item \textbf{PLLuM-12B}\footnote{\fontsize{7}{9}{\url{https://huggingface.co/CYFRAGOVPL/PLLuM-12B-nc-chat}}} \cite{pllum2025} derived from Mistral-Nemo-Base-2407.
\end{enumerate}

\paragraph{Reference translation model}

 For translation experiments, \textsl{NLLB-200-distilled-1.3B}\footnote{\fontsize{7}{9}{\url{https://huggingface.co/facebook/nllb-200-distilled-1.3B}}} \cite{nllbteam2022languageleftbehindscaling} is used as the state-of-the-art reference model.

 \paragraph{LLM-as-a-judge evaluation model}

 For LLM-as-a-judge evaluation, \textsl{Qwen/Qwen3-30B-A3B-Thinking-2507}\footnote{\fontsize{7}{9}{\url{https://huggingface.co/Qwen/Qwen3-30B-A3B-Thinking-2507}}} \cite{qwen3technicalreport} is used with the following hyper-parameters: \texttt{temperature} of 0.6, \texttt{top\_p} of 0.95 and \texttt{top\_k} of 20.

\newpage
\section{Instruction tuning hyper-parameters}
\label{sec:appendixB}
\begin{table}[h]
\centering
\begin{tabular}{l|l}
\toprule
\multicolumn{2}{c}{\textbf{LoRA configuration}} \\
\midrule
Attention implementation & Flash attention 2 \\
LoRA rank ($r$) & 128 \\
LoRA alpha ($\alpha$) & 256 \\
LoRA dropout & 0.05 \\
\midrule
\multicolumn{2}{c}{\textbf{SFT configuration}} \\
\midrule
Number of training epochs & 3 \\
Per device train batch size & 2 \\
Per device eval batch size & 2 \\
Gradient accumulation steps & 4 \\
Learning rate & 2e-5 \\
Weight decay & 1e-3 \\
Adam beta1 & 0.999 \\
Adam beta2 & 0.9 \\
Warmup ratio & 0.05 \\
Learning rate scheduler & cosine \\
Max sequence length & 8192\ /\ 4096 \\
\bottomrule
\end{tabular}
\caption{\label{tab:hyperparams} LoRA and SFT hyperparameters. For Bielik-7B, the maximum sequence length is set to 4096, its maximum context window; all other models use 8192.}
\end{table}

\paragraph{Frameworks and libraries used in training:}
\begin{itemize}
    \item DeepSpeed: \url{https://github.com/microsoft/DeepSpeed}, 
    \item FlashAttention \cite{dao2022flashattentionfastmemoryefficientexact}: \url{https://github.com/Dao-AILab/flash-attention},
    \item HuggingFace's ecosystem:
    \begin{itemize}
        \item transformers: \url{https://github.com/huggingface/transformers},
        \item datasets: \url{https://github.com/huggingface/datasets},
        \item trl: \url{https://github.com/huggingface/trl},
        \item peft: \url{https://github.com/huggingface/peft}
    \end{itemize}
    \item LoRA \cite{hu2021loralowrankadaptationlarge}.
\end{itemize}

\section{Normalisation process}
\label{sec:appendixC}
Gender-inclusive proofreading involves automatically transforming a text that may contain gender-biased expressions (\ref{item:normalisation:1}) into a gender-inclusive version, e.g. \ref{item:possible:prediction:h} in \ref{item:normalisation:2}. When proofreading the input example, LLMs may generate any of the gender-inclusive candidates listed in \ref{item:normalisation:2}. Therefore, gender-inclusive proofreading can be treated as an open-ended task.

Nevertheless, our normalisation process (\ref{item:normalisation:3}) allows us to transform all possible predictions returned by the LLM, i.e.  \ref{item:possible:prediction:a}--\ref{item:possible:prediction:h}, into a unified token list (see the normalised tokens in Table \ref{tab:adx:normalisation_example}). The normalisation process involves expanding gender-inclusive expressions, especially those containing a gender star, into their full masculine and feminine forms. We then remove predefined stop words (i.e. conjunctions and punctuation marks) and convert all tokens to lowercase. The resulting normalised output (\texttt{generated\_target}) is stored as the value of the \texttt{normalised\_target} field in the JSON structure (see Figure \ref{fig:apx:running_example_gold}).

\begin{enumerate}
\item\label{item:normalisation:1} \textbf{LLM Input}

\parbox[t]{0.95\linewidth}{\textcolor{NavyBlue}{Tegoroczni laureaci} Oscarów \textcolor{Orange}{pozowali} na czerwonym dywanie.\\[0.3ex]
[gloss.] \textcolor{NavyBlue}{This year's-ADJ.$_{\textit masc}$} \textcolor{NavyBlue}{winners$_{\textit masc}$} of Oscars \textcolor{Orange}{posed$_{\textit masc}$} on the red carpet\\[0.3ex]
{\fontsize{9.5}{10}\selectfont [Eng.] \textcolor{NavyBlue}{This year's} Oscar \textcolor{NavyBlue}{winners} \textcolor{Orange}{posed} on the red carpet.}}

\item\label{item:normalisation:2}\textbf{Possible LLM gender-inclusive predictions}

\begin{enumerate}[label=\alph*)]
\item\label{item:possible:prediction:a} \textcolor{NavyBlue}{Tegoroczni laureaci} i \textcolor{Green}{tegoroczne laureatki} Oscarów \textcolor{Orange}{pozowali} i \textcolor{Red}{pozowały} na czerwonym dywanie.
\item \textcolor{Green}{Tegoroczne laureatki} i \textcolor{NavyBlue}{tegoroczni laureaci} Oscarów \textcolor{Red}{pozowały} i  \textcolor{Orange}{pozowali} na czerwonym dywanie.
\item \textcolor{NavyBlue}{Tegoroczni laureaci} i \textcolor{Green}{tegoroczne laureatki} Oscarów \textcolor{Orange}{pozowali}/\textcolor{Red}{pozowały} na czerwonym dywanie.
\item \textcolor{NavyBlue}{Tegoroczni laureaci}/\textcolor{Green}{Tegoroczne laureatki} Oscarów \textcolor{Orange}{pozowali}/\textcolor{Red}{pozowały} na czerwonym dywanie.
\item \textcolor{Green}{Tegoroczne laureatki}/\textcolor{NavyBlue}{Tegoroczni laureaci} Oscarów \textcolor{Red}{pozowały}/\textcolor{Orange}{pozowali} na czerwonym dywanie.
\item \textcolor{NavyBlue}{Tegoroczni}/\textcolor{Green}{Tegoroczne} \textcolor{NavyBlue}{laureaci}/\textcolor{Green}{laureatki} Oscarów \textcolor{Orange}{pozowali}/\textcolor{Red}{pozowały} na czerwonym dywanie.
\item\label{item:possible:prediction:g} \textcolor{NavyBlue}{Tegoroczni}/\textcolor{Green}{Tegoroczne} \textcolor{NavyBlue}{laureaci}/\textcolor{Green}{laureatki} Oscarów pozowa*\textcolor{Orange}{li}/\textcolor{Red}{ły} na czerwonym dywanie.
\item\label{item:possible:prediction:h} Tegoroczn*\textcolor{NavyBlue}{i}/\textcolor{Green}{e} laurea*\textcolor{NavyBlue}{ci}/\textcolor{Green}{tki} Oscarów pozowa*\textcolor{Orange}{li}/\textcolor{Red}{ły} na czerwonym dywanie.
\end{enumerate}
 
\item\label{item:normalisation:3}\textbf{Normalisation}

\begin{table}[h!]
\centering
\renewcommand\tabcolsep{18pt}
\renewcommand*{\arraystretch}{0.9}
\begin{tabular}{ll}
\toprule
Polish & English glosses \\
\midrule
\texttt{\textcolor{NavyBlue}{tegoroczni}} & \texttt{\textcolor{NavyBlue}{this year's$_{masc}$}} \\
\texttt{\textcolor{NavyBlue}{laureaci}} & \texttt{\textcolor{NavyBlue}{winners$_{masc}$}} \\
\texttt{\textcolor{Green}{tegoroczne}} & \texttt{\textcolor{Green}{this year's$_{fem}$}} \\
\texttt{\textcolor{Green}{laureatki}} & \texttt{\textcolor{Green}{winners$_{fem}$}} \\
\texttt{oscarów} & \texttt{oscar} \\
\texttt{\textcolor{Orange}{pozowali}} & \texttt{\textcolor{Orange}{posed$_{masc}$}} \\
\texttt{\textcolor{Red}{pozowały}} & \texttt{\textcolor{Red}{posed$_{fem}$}} \\
\texttt{na} & \texttt{on} \\
\texttt{czerwonym} & \texttt{red} \\
\texttt{dywanie} & \texttt{carpet} \\
\bottomrule
\end{tabular}
\caption{\label{tab:adx:normalisation_example} Normalisation of all possible gendered dyads in the running example. Colour legend: \textcolor{NavyBlue}{masculine NPs}, \textcolor{Green}{feminine NPs},  \textcolor{Orange}{masculine verb}, and \textcolor{Red}{feminine verb}. All tokens are lowercased. Punctuation marks and coordinating conjunction \textit{i} ['and'] are filtered out.}
\end{table}

\item\label{item:normalisation:4}\textbf{Normalisation output}

\begin{figure}[h!]
\begin{minipage}{\textwidth}
{\small
\begin{verbatim}
{'source': 'Tegoroczni laureaci Oskarów pozowali
    na czerwonym dywanie.',
 'generated_target': 'Tegoroczn*i/e laurea*ci/tki
    Oscarów pozowa*li/ły na czerwonym dywanie.',
 'normalised_target': ['tegoroczni',
    'tegoroczne','laureaci', 'laureatki',
    'oscarów', 'pozowali','pozowały', 'na',
    'czerwonym', 'dywanie']}
\end{verbatim}
}
\end{minipage}
\caption{The JSON dictionary of the running example.}
\label{fig:apx:running_example_gold}
\end{figure}
\end{enumerate}

\section{Detailed results}
\label{sec:appendixD}

For the proofreading task, we report the main evaluation metrics -- precision, recall, and F$_1$ -- on which our conclusions are based, and text similarity metrics -- BLEU, chrF, and chrF++ -- to assess the fluency of proofread texts (Table \ref{tab:proofreading_D}). We do not report accuracy, because it is less informative; label distribution is highly imbalanced in the proofreading dataset (most tokens do not require modification) and token-level accuracy would be dominated by correctly predicting unchanged tokens.

For the translation task, we report text similarity metrics -- BLEU, chrF, and chrF++ (Table \ref{tab:translation_D}).

{%
\onecolumn
\centering
\renewcommand*{\arraystretch}{1.3}
\renewcommand\tabcolsep{12pt}
\small
\begin{longtable}{l||ccc|ccc}
    \toprule
        {\textbf{LLM}}  & {\textbf{Precision}} & {\textbf{Recall}} & {\textbf{F$_1$}} 
        & \textsc{\textbf{bleu}} & {\textbf{chrF}} & {\textbf{chrF++}} \\
\midrule
\multicolumn{7}{c}{Default LLMs}\\
\midrule
llama-8b-default & 0.19 & 0.54 & 0.28 & 47.44 & 75.38 & 73.91 \\
llama-8b-default-pl & 0.35 & 4.12 & 0.64 & 41.94 & 76.24 & 75.72\\
llama-8b-default-en & 0.29 & \underline{7.04} & 0.56 & 36.13 & 74.76 & 74.12 \\
\midrule
mistral-7b-default & 0.08 & 0.19 & 0.11 & 14.09 & 43.32 & 40.53 \\
mistral-7b-default-pl & 0.12 & 2.01 & 0.22 & 9.20 & 40.50 & 38.87\\
mistral-7b-default-en & 0.13 & 1.20 & 0.24 & 14.43 & 48.43 & 46.55 \\
\midrule
mistral-nemo-12b-default & 0.42 & 0.47 & 0.44 & 62.98 & 78.33 & 76.95 \\
mistral-nemo-12b-default-pl & 1.30 & 6.14 & 2.15 & 61.99 & \textbf{86.25} & \textbf{85.67}\\
mistral-nemo-12b-default-en & 0.64 & 8.32 & 1.19 & 36.72 & 76.36 & 75.02 \\
\midrule
\midrule
llama-pllum-8b-default & 0.25 & 0.30 & 0.27 & 32.04 & 57.13 & 54.36 \\
llama-pllum-8b-default-pl & \underline{0.50} & 0.48 & 0.49 & 41.94 & 51.25 & 50.05\\
llama-pllum-8b-default-en & 0.42 & 1.16 & 0.62 & 40.69 & 67.13 & 65.83 \\
\midrule
bielik-7b-default & 0.46 & 0.59 & 0.52 & \underline{\textbf{66.17}} & \underline{80.40} & 79.39 \\
bielik-7b-default-pl & 0.36 & 2.88 & \underline{0.64} & 51.71 & \underline{80.40} & \underline{80.04}\\
bielik-7b-default-en & 0.24 & 2.70 & 0.45 & 39.99 & 74.61 & 74.04 \\
\midrule
pllum-12b-default & 1.25 & 1.87 & 1.50 & 41.65 & 67.70 & 65.26 \\
pllum-12b-default-pl & \textbf{2.77} & 6.90 & \textbf{3.95} & 63.59 & 82.74 & 81.86\\
pllum-12b-default-en & 2.22 & 6.43 & 3.30 & 58.71 & 81.13 & 80.17 \\
\midrule
bielik-11b-default & 0.40 & 0.74 & 0.52 & 39.12 & 66.54 & 64.18 \\
bielik-11b-default-pl & 1.57 & 10.46 & 2.74 & 56.56 & 83.93 & 83.22\\
bielik-11b-default-en & 1.73 & \textbf{15.34} & 3.11 & 55.79 & 84.72 & 84.08 \\
\midrule
\multicolumn{7}{c}{Few-shot LLMs}\\
\midrule
llama-8b-fewshot & 0.41 & 1.98 & 0.68 & 51.61 & 80.06 & 79.18 \\
llama-8b-fewshot-pl & 0.57 & \underline{5.78} & \underline{1.04} & \underline{51.91} & \underline{82.57} & \underline{82.04}\\
llama-8b-fewshot-en & 0.39 & 6.13 & 0.72 & 44.35 & 79.61 & 78.99 \\
\midrule
mistral-7b-fewshot & 0.13 & 0.69 & 0.22 & 15.13 & 48.39 & 46.21 \\
mistral-7b-fewshot-pl & 0.28 & 2.69 & 0.50 & 18.81 & 56.24 & 54.55\\
mistral-7b-fewshot-en & 0.28 & 1.35 & 0.47 & 21.01 & 54.94 & 52.94 \\
\midrule
mistral-nemo-12b-fewshot & 0.84 & 1.90 & 1.16 & \textbf{68.43} &  \textbf{84.95} &  \textbf{84.19} \\
mistral-nemo-12b-fewshot-pl & 0.86 & 2.39 & 1.26 & 50.53 & 74.39 & 73.41\\
mistral-nemo-12b-fewshot-en & 0.50 & 1.94 & 0.79 & 34.29 & 65.10 & 63.78 \\
\midrule
\midrule
llama-pllum-8b-fewshot & 0.54 & 0.64 & 0.59 & 37.51 & 52.38 & 50.80 \\
llama-pllum-8b-fewshot-pl & \underline{0.60} & 0.71 & 0.65 & 46.65 & 58.55 & 57.44\\
llama-pllum-8b-fewshot-en & 0.50 & 0.67 & 0.58 & 43.77 & 58.28 & 57.14 \\
\midrule
bielik-7b-fewshot\footnote{The context size of the bielik-7b model makes it impossible to carry out this experiment.} & NA & NA & NA & NA & NA & NA \\
\midrule
pllum-12b-fewshot & 1.89 & 2.76 & 2.25 & 50.84 & 68.61 & 67.15 \\
pllum-12b-fewshot-pl & \textbf{2.47} & 2.86 & \textbf{2.65} & 58.62 & 69.98 & 68.83\\
pllum-12b-fewshot-en & 2.17 & 2.57 & 2.36 & 56.50 & 68.30 & 67.17 \\
\midrule
bielik-11b-fewshot & 1.20 & 5.10 & 1.94 & 52.91 & 80.23 & 79.12 \\
bielik-11b-fewshot-pl & 1.43 & \textbf{13.10} & 2.58 & 54.92 & 84.01 & 83.39\\
bielik-11b-fewshot-en & 1.41 & 9.71 & 2.47 & 53.69 & 82.93 & 82.09 \\
\midrule
        {\textbf{LLM}} & {\textbf{Precision}} & {\textbf{Recall}} & {\textbf{F$_1$}}
        & \textsc{\textbf{bleu}} & {\textbf{chrF}} & {\textbf{chrF++}} \\
\midrule
\multicolumn{7}{c}{IPIS-tuned LLMs}\\
\midrule
llama-8b-tuned & 64.34 & 47.15 & 54.42 & 93.92 & 97.47 & 97.29 \\
llama-8b-tuned-pl & 53.43 & 41.58 & 46.76 & 92.85 & 96.62 & 96.46\\
llama-8b-tuned-en & 55.94 & 44.00 & 49.26 & 93.47 & 97.01 & 96.80 \\
\midrule
mistral-7b-tuned & 43.78 & 54.70 & 48.64 & 93.54 & 97.25 & 97.04 \\
mistral-7b-tuned-pl & 27.11 & 48.50 & 34.78 & 91.29 & 96.50 & 96.27\\
mistral-7b-tuned-en & 31.87 & 49.44 & 38.76 & 90.53 & 96.43 & 96.20 \\
\midrule
mistral-nemo-12b-tuned & 64.82 & 56.24 & 60.23 & 94.45 & 97.72 & 97.51 \\
mistral-nemo-12b-tuned-pl & 56.93 & 51.90 & 54.30 & 93.28 & 97.22 & 97.03\\
mistral-nemo-12b-tuned-en & 50.70 & 51.68 & 51.19 & 94.09 & 97.33 & 97.13 \\
\midrule
\midrule
llama-pllum-8b-tuned & \underline{69.72} & 52.79 & 60.08 & 94.28 & 97.64 & 97.48 \\
llama-pllum-8b-tuned-pl & 57.36 & 41.48 & 48.15 & 93.40 & 96.85 & 96.69\\
llama-pllum-8b-tuned-en & 57.89 & 50.15 & 53.75 & 93.78 & 97.25 & 97.05 \\
\midrule
bielik-7b-tuned & 61.25 & \underline{59.93} & \underline{60.58} & \underline{94.67} & \underline{97.71} & \underline{97.49} \\
bielik-7b-tuned-pl & 59.59 & 52.16 & 55.63 & 94.10 & 97.30 & 97.09\\
bielik-7b-tuned-en & 56.68 & 54.50 & 55.57 & 93.62 & 97.26 & 97.03 \\
\midrule
pllum-12b-tuned & 70.82 & 63.19 & 66.79 & 95.10 & 97.95 & 97.76 \\
pllum-12b-tuned-pl & 67.92 & 57.41 & 62.23 & 94.52 & 97.68 & 97.50\\
pllum-12b-tuned-en & 69.63 & 60.17 & 64.55 & 94.68 & 97.77 & 97.58 \\
\midrule
bielik-11b-tuned & \textbox{\textbf{72.68}} & \textbox{\textbf{65.09}} & \textbox{\textbf{68.68}} & \textbox{\textbf{95.22}} & \textbox{\textbf{97.99}} & \textbox{\textbf{97.81}} \\
bielik-11b-tuned-pl & 32.09 & 57.58 & 41.22 & 91.82 & 96.93 & 96.74\\
bielik-11b-tuned-en & 58.67 & 59.24 & 58.95 & 94.82 & 97.61 & 97.43 \\
\bottomrule
\caption{Evaluation of multilingual and Polish-specific LLMs in solving the \textbf{gender-inclusive proofreading} task. Explanation: \textbf{default} -- the default LLM, \textbf{fewshot} -- the default LLM in the few-shot setting, \textbf{tuned} -- the LLM tuned on IPIS proofreading instructions; \textbf{pl} -- a system prompt in Polish; \textbf{en} -- a system prompt in English; \textbf{bold} -- the best model within a scenario (\textsl{default}, \textsl{fewshot}, and IPIS-\textsl{tuned}); \underline{underline} -- the best small-size model; \textbox{framed} -- the best model overall.}
\label{tab:proofreading_D}
 \end{longtable}

\newpage
\renewcommand*{\arraystretch}{1.32}
\renewcommand\tabcolsep{.7pt}
\small
\begin{longtable}{l||ccc|ccc||ccc|ccc}
    \toprule
        \multirow{3}{*}{\textbf{LLM}} & \multicolumn{6}{c||}{\textbf{Polish$\rightarrow$English}} & \multicolumn{6}{c}{\textbf{English$\rightarrow$Polish}}\\
         \cmidrule(lr){2-7}  \cmidrule(lr){8-13}
         & \multicolumn{3}{c|}{\textbf{PL user prompt}} & \multicolumn{3}{c||}{\textbf{EN user prompt}} & \multicolumn{3}{c|}{\textbf{PL user prompt}} & \multicolumn{3}{c}{\textbf{EN user prompt}}\\
        & \textsc{\textbf{bleu}} & \textbf{chrF} & \textbf{chrF++} & 
        \textsc{\textbf{bleu}} & \textbf{chrF} & \textbf{chrF++} &
       \textsc{\textbf{bleu}} & \textbf{chrF} & \textbf{chrF++} &
        \textsc{\textbf{bleu}} & \textbf{chrF} & \textbf{chrF++} \\
            \midrule
\multicolumn{13}{c}{Default LLMs}\\
\midrule
llama-8b-default & 24.69 & 42.79 & 40.68 & 30.36 & 56.19 & 54.04 & 26.84 & 62.85 & 59.06 & 24.69 & 61.93 & 58.08\\
llama-8b-default-pl & 22.72 & 55.62 & 54.02 & 21.92 & 58.19 & 56.48 & 16.90 & 57.14 & 53.37 & 16.84 & 57.66 & 53.86\\
llama-8b-default-en & 26.76 & 60.78 & 58.88 & 26.87 & 65.11 & 63.18 & 13.81 & 55.41 & 51.57 & 9.89 & 49.63 & 45.95\\
\midrule
mistral-7b-default & 41.97 & 68.02 & 65.91 & \underline{42.48} & \underline{69.81} & \underline{67.56} & 6.99 & 44.32 & 40.50 & 9.91 & 50.21 & 45.92\\
mistral-7b-default-pl & 18.05 & 55.24 & 53.20 & 15.55 & 53.88 & 51.91 & 3.59 & 34.57 & 31.38 & 2.93 & 30.92 & 27.95\\
mistral-7b-default-en & 18.90 & 56.84 & 54.90 & 24.18 & 60.24 & 58.19 & 4.48 & 38.65 & 35.16 & 4.28 & 37.23 & 33.75\\
\midrule
mistral-nemo-12b-default & \textbf{53.68} & \textbf{75.35} & \textbf{73.57} & \textbf{53.42} & \textbf{75.78} & \textbf{73.97} & 23.75 & 60.16 & 56.33 & 23.11 & 59.66 & 55.63\\
mistral-nemo-12b-default-pl & 42.62 & 71.94 & 70.07 & 47.29 & 74.17 & 72.32 & 14.75 & 54.89 & 51.00 & 12.23 & 53.75 & 49.71\\
mistral-nemo-12b-default-en & 40.79 & 72.66 & 70.86 & 40.26 & 72.18 & 70.23 & 10.15 & 49.99 & 46.11 & 11.06 & 53.03 & 48.76\\
\midrule
\midrule
llama-pllum-8b-default & \underline{42.85} & \underline{69.52} & \underline{67.72} & 40.53 & 67.65 & 65.89 & \underline{31.14} & \underline{64.95} & \underline{61.47} & \underline{\textbf{34.28}} & \underline{66.86} & \underline{63.56}\\
llama-pllum-8b-default-pl & 32.55 & 51.67 & 50.17 & 30.19 & 50.42 & 48.87 & 18.39 & 42.56 & 39.97 & 15.87 & 43.12 & 40.30\\
llama-pllum-8b-default-en & 33.15 & 65.33 & 63.44 & 27.21 & 61.81 & 60.12 & 25.40 & 59.01 & 55.72 & 24.79 & 62.76 & 59.39\\
\midrule
bielik-7b-default & 38.98 & 66.02 & 63.73 & 36.30 & 61.78 & 59.67 & 28.10 & 64.36 & 60.53 & 27.61 & 63.44 & 59.71\\
bielik-7b-default-pl & 35.14 & 60.53 & 58.47 & 39.08 & 64.78 & 62.68 & 25.62 & 61.03 & 57.32 & 27.01 & 62.38 & 58.77\\
bielik-7b-default-en & 36.63 & 64.63 & 62.45 & 36.93 & 62.81 & 60.78 & 20.02 & 58.47 & 54.72 & 22.46 & 60.44 & 56.83\\
\midrule
pllum-12b-default & 43.51 & 71.09 & 69.17 & 44.34 & 71.24 & 69.30 & 32.09 & 66.47 & 63.14 & 33.17 & 66.04 & 62.70\\
pllum-12b-default-pl & 38.27 & 66.96 & 64.88 & 38.87 & 67.72 & 65.61 & 27.90 & 60.39 & 57.01 & 29.11 & 63.01 & 59.59\\
pllum-12b-default-en & 41.52 & 68.24 & 66.24 & 36.18 & 65.66 & 63.64 & 25.70 & 57.50 & 54.33 & 24.91 & 61.32 & 57.96\\
\midrule
bielik-11b-default & 47.60 & 73.39 & 71.45 & 47.54 & 72.50 & 70.59 & \textbf{41.49} & \textbf{71.78} & \textbf{68.79} & 27.39 &65.30 & 62.49\\
bielik-11b-default-pl & 46.78 & 73.08 & 71.16 & 43.67 & 70.21 & 68.17 & 35.80 & 69.77 & 66.65 & 32.31 & \textbf{68.94} & \textbf{65.86}\\
bielik-11b-default-en & 47.99 & 73.76 & 71.78 & 36.39 & 68.39 & 66.52 & 32.70 & 68.65 & 65.67 & 32.13 & 68.54 & 65.51\\
\midrule
\multicolumn{13}{c}{Few-shot LLMs}\\
\midrule
llama-8b-fewshot & \underline{43.03} & 65.99 & 64.07 & \underline{44.27} & \underline{71.15} & \underline{69.08} & \underline{26.37} & \underline{62.10} & \underline{58.24} & 25.31 & \underline{61.92} & \underline{58.09}\\
llama-8b-fewshot-pl & 37.46 & 65.93 & 64.05 & 34.21 & 65.63 & 63.80 & 18.32 & 58.38 & 54.54 & 18.24 & 58.75 & 54.93\\
llama-8b-fewshot-en & 29.26 & 58.95 & 57.04 & 29.22 & 65.03 & 63.04 & 15.49 & 57.01 & 53.09 & 10.29 & 51.07 & 47.29\\
\midrule
mistral-7b-fewshot & 40.94 & \underline{68.92} & \underline{66.59} & 40.10 & 68.22 & 65.83 & 8.27 & 47.02 & 42.98 & 9.35 & 48.28 & 44.22\\
mistral-7b-fewshot-pl & 31.11 & 62.88 & 60.77 & 30.83 & 64.28 & 62.05 & 7.24 & 46.29 & 42.26 & 4.90 & 39.57 & 35.99\\
mistral-7b-fewshot-en & 32.99 & 65.67 & 63.44 & 39.98 & 68.22 & 65.79 & 4.69 & 38.35 & 34.85 & 9.72 & 49.43 & 45.22\\
\midrule
mistral-nemo-12b-fewshot & \textbf{54.52} & \textbf{75.89} & \textbf{74.15} & \textbf{51.78} & \textbf{74.56} & \textbf{72.74} & 20.08 & 53.94 & 50.23 & 17.56 & 52.80 & 49.08\\
mistral-nemo-12b-fewshot-pl & 29.65 & 56.87 & 55.03 & 41.43 & 70.16 & 68.40 & 14.67 & 50.99 & 47.12 & 12.03 & 50.80 & 46.80\\
mistral-nemo-12b-fewshot-en & 32.20 & 66.12 & 64.40 & 37.20 & 70.33 & 68.45 & 8.35 & 45.42 & 41.53 & 11.08 & 51.95 & 47.82\\
\midrule
\midrule
llama-pllum-8b-fewshot & 32.00 & 59.06 & 57.38 & 33.73 & 61.88 & 59.97 & 21.75 & 51.89 & 48.60 & \underline{25.75} & 60.75 & 57.31\\
llama-pllum-8b-fewshot-pl & 32.74 & 50.02 & 48.51 & 21.43 & 35.63 & 34.37 & 15.38 & 34.91 & 32.67 & 9.87 & 29.05 & 26.74\\
llama-pllum-8b-fewshot-en & 33.16 & 64.09 & 62.32 & 30.87 & 63.08 & 61.43 & 19.06 & 50.65 & 47.46 & 19.91 & 59.51 & 56.23\\
\midrule
pllum-12b-fewshot 		& 28.67 & 48.68 & 46.72 & 33.39 & 58.32 & 56.18 & 17.55 & 41.36 & 38.59 & 25.37 & 55.91 & 52.73\\
pllum-12b-fewshot-pl 	& 36.29 & 59.81 & 57.95 & 37.94 & 65.75 & 63.58 & 12.57 & 33.94 & 31.45 & 30.20 & 62.12 & 58.81\\
pllum-12b-fewshot-en 	& 34.90 & 63.21 & 61.21 & 35.73 & 61.04 & 58.92 & 19.23 & 43.92&41.00& 23.11 & 56.56 & 53.32\\
\midrule
bielik-11b-fewshot & 50.01 & 73.93 & 72.04 & 49.66 & 73.99 & 72.04 & 38.63 & 68.78 & 66.09 & 33.19 & 68.11 & 65.34\\
bielik-11b-fewshot-pl & 49.38 & 73.84 & 71.90 & 49.55 & 74.02 & 72.03 & 37.81 & 69.92 & 67.06 & 42.76 & 72.19 & 69.32\\
bielik-11b-fewshot-en & 48.33 & 73.43 & 71.48 & 49.14 & 73.75 & 71.77 & \textbox{\textbf{43.02}} & \textbox{\textbf{72.46}} &  \textbox{\textbf{69.61}} &  \textbox{\textbf{43.17}} &  \textbox{\textbf{72.82}} &  \textbox{\textbf{70.00}}\\
\midrule
        \multirow{3}{*}{\textbf{LLM}} & \multicolumn{6}{c||}{\textbf{Polish$\rightarrow$English}} & \multicolumn{6}{c}{\textbf{English$\rightarrow$Polish}}\\
         \cmidrule(lr){2-7}  \cmidrule(lr){8-13}
         & \multicolumn{3}{c|}{\textbf{PL user prompt}} & \multicolumn{3}{c||}{\textbf{EN user prompt}} & \multicolumn{3}{c|}{\textbf{PL user prompt}} & \multicolumn{3}{c}{\textbf{EN user prompt}}\\
        & \textsc{\textbf{bleu}} & \textbf{chrF} & \textbf{chrF++} & 
        \textsc{\textbf{bleu}} & \textbf{chrF} & \textbf{chrF++} &
        \textsc{\textbf{bleu}} & \textbf{chrF} & \textbf{chrF++} &
        \textsc{\textbf{bleu}} & \textbf{chrF} & \textbf{chrF++} \\
\midrule
\multicolumn{13}{c}{IPIS-tuned LLMs}\\
\midrule
llama-8b-tuned & 29.45 & 61.48 & 60.08 & 30.82 & 62.02 & 60.55 & 34.15 & 65.31 & 62.30 & 33.93 & 65.82 & 62.77\\
llama-8b-tuned-pl & 31.72 & 61.64 & 60.11 & 41.62 & 63.64 & 62.19 & 31.55 & 62.38 & 59.40 & 31.18 & 62.25 & 59.28\\
llama-8b-tuned-en & 40.23 & 64.46 & 62.83 & 43.17 & 65.39 & 63.89 & 30.48 & 61.57 & 58.45 & 29.07 & 60.03 & 56.98\\
\midrule
mistral-7b-tuned & 10.43 & 39.85 & 39.15 & 12.93 & 44.39 & 43.65 & 26.10 & 59.90 & 56.69 & 25.51 & 57.32 & 54.43\\
mistral-7b-tuned-pl & 24.13 & 58.52 & 57.53 & 22.57 & 57.04 & 56.16 & 17.77 & 52.04 & 49.15 & 22.70 & 57.70 & 54.42\\
mistral-7b-tuned-en & 16.08 & 49.89 & 48.86 & 19.11 & 53.28 & 52.27 & 24.49 & 55.35 & 52.50 & 24.85 & 57.26 & 54.24\\
\midrule
mistral-nemo-12b-tuned & 10.75 & 39.89 & 39.25 & 16.12 & 49.01 & 48.30 & 26.35 & 60.41 & 57.73 & 34.17 & 65.99 & 62.85\\
mistral-nemo-12b-tuned-pl & 14.25 & 47.17 & 46.28 & 21.04 & 56.65 & 55.76 & 21.66 & 56.67 & 53.71 & 22.05 & 57.95 & 54.97\\
mistral-nemo-12b-tuned-en & 14.12 & 46.21 & 45.58 & 10.69 & 39.92 & 39.27 & 19.00 & 55.48 & 52.59 & 25.07 & 59.97 & 56.90\\
\midrule
llama-pllum-8b-tuned & 42.83 & 71.54 & 70.17 & 48.04 & \underline{73.88} & \underline{72.43} & 37.65 & 67.65 & 64.87 & 36.66 & 66.61 & \underline{\textbf{63.86}}\\
llama-pllum-8b-tuned-pl & 48.94 & \underline{72.00} & \underline{70.54} & 48.86 & 72.50 & 71.05 & \underline{\textbf{39.21}} & \underline{\textbf{68.23}} & \underline{\textbf{65.34}} & \underline{\textbf{38.30}} & \underline{\textbf{66.56}} & 63.68\\
llama-pllum-8b-tuned-en & 47.28 & 71.16 & 69.63 & 44.66 & 68.98 & 67.39 & 34.84 & 66.46 & 63.63 & 32.12 & 65.33 & 62.46\\
\midrule
\midrule
bielik-7b-tuned & 49.11 & 71.86 & 70.33 & 48.83 & 71.68 & 70.17 & 34.64 & 64.77 & 61.95 & 32.15 & 62.48 & 59.58\\
bielik-7b-tuned-pl & 48.56 & 71.46 & 69.93 & 46.91 & 69.80 & 68.27 & 36.79 & 64.81 & 61.82 & 36.49 & 64.60 & 61.56\\
bielik-7b-tuned-en & 39.45 & 66.51 & 65.03 & 42.18 & 67.61 & 66.15 & 37.43 & 64.82 & 61.91 & 36.86 & 65.08 & 62.11\\
\midrule
pllum-12b-tuned & 54.98 & 76.95 & 75.53 & 52.98 & 75.58 & 74.20 & 27.49 & 63.16 & 60.63 & 33.03 & 65.35 & 62.65\\
pllum-12b-tuned-pl & 48.23 & 73.02 & 71.56 & 44.93 & 71.35 & 69.90 & 24.69 & 60.32 & 57.32 & 29.74 & 63.86 & 60.91\\
pllum-12b-tuned-en & 47.12 & 72.90 & 71.31 & 47.24 & 72.92 & 71.39 & 23.33 & 59.05 & 56.35 & 21.89 & 58.70 & 55.85\\
\midrule
bielik-11b-tuned & 55.19 & 75.80 & 74.40 & \textbox{\textbf{57.45}} & \textbox{\textbf{77.93}} & \textbox{\textbf{76.52}} & 34.26 & 55.08 & 52.63 & 35.04 & 55.92 & 53.44\\
bielik-11b-tuned-pl & 56.70 & 76.93 & 75.54 & 55.24 & 75.35 & 73.90 & 31.70 & 58.36 & 55.62 & 26.74 & 55.96 & 53.25\\
bielik-11b-tuned-en & \textbox{\textbf{57.55}} & \textbox{\textbf{78.03}} & \textbox{\textbf{76.66}} & 57.30 & 76.63 & 75.29 & 28.71 & 60.97 & 58.12 & 25.96 & 58.93 & 55.77\\
\midrule
\midrule
 \multicolumn{13}{c}{\textbf{standarised human-annotated IPIS-translation-test}}\\
 \midrule
nllb-200-distilled-1.3b & 45.47$^*$ &  65.59$^*$ & 66.04$^*$ & -- & -- & -- & \textbf{\textcolor{Green}{38.55}}$^*$ & 65.34$^*$ & 65.14$^*$ & -- & -- & --\\
bielik-11b-default & 50.58 & 74.12 & 74.52 & 47.61 &  71.11 & 71.41 & 34.62 &  69.46 & 69.27 & 32.89 & 68.67 & 68.45 \\
    gemma-3-27b & 48.70 & 78.41 & 78.58 & \textbf{\textcolor{Green}{53.81}} & \textbf{\textcolor{Green}{79.16}} & \textbf{\textcolor{Green}{79.39}} & 37.50 & \textbf{\textcolor{Green}{71.22}} & \textbf{\textcolor{Green}{70.65}} & 13.95 & 60.12 & 58.98\\
\bottomrule
\caption{\label{tab:translation_D}Preliminary evaluation of multilingual and Polish-specific LLMs on the \textbf{gender-sensitive translation} task using the human-annotated IPIS-translation-test subset. Explanation: \textbf{tuned} -- LLM tuned on IPIS translation instructions; \textbf{pl} -- a system prompt in Polish; \textbf{en} -- a system prompt in English; \textbf{bold} -- the best model within a scenario (\textsl{default}, \textsl{fewshot}, and IPIS-\textsl{tuned}); \underline{underline} -- the best small-size model; \textbox{framed} -- the best model overall.\\
$[$Lower section of the result table$]$ Reference Polish--English translation quality is reported for the out-of-the-shelf \textsl{NLLB-200-distilled-1.3B}, \textsl{Bielik-11B} and Gemma-3-27B models. The models are evaluated on the \textbf{standardised} human-annotated IPIS-translation-test comprising standard Polish--standard English translation pairs. Explanations: $^*$user prompts are not used; \textbf{\textcolor{Green}{highest performance on standard translation}}.}
 \end{longtable}
}
\twocolumn

\section{Human vs. LLM evaluation of English-to-Polish translation}
\label{sec:appendix_llm_human}
We compare the scoring behaviour of the human evaluator and the LLM judge across the predefined evaluation criteria -- MT-related criteria (1--4) and gender inclusivity criterion (5): 
\begin{enumerate}
\setlength\itemsep{-.2em}
    \item \textsl{completeness}, 
    \item \textsl{clarity} (and logical coherence), 
    \item \textsl{structure} (i.e. structural fidelity),
    \item (linguistic) \textsl{correctness},
    \item \textsl{gender inclusivity}.
\end{enumerate}

\subsection{MT-related evaluation}
Both evaluators assign higher scores to the high-\textsc{bleu} subset of translations generated by IPIS-tuned Bielik-11B than to the low-\textsc{bleu} ones (see Table \ref{tab:llm-as-a-judge-table}). For the high-\textsc{bleu} instances, the human evaluator assigns high scores across all MT-related criteria ($4.053-4.965$). For the low-\textsc{bleu} subset, \textsl{completeness} ($1.333$) and \textsl{structure} ($3.018$) are clearly lower scored, but \textsl{clarity} ($4.333$) and \textsl{correctness} ($4.509$) remain relatively high, i.e. translations with low \textsc{bleu} scores can still be linguistically well-formed and understandable.

The LLM judge shows more uniform scoring patterns across MT-related criteria. For the high-\textsc{bleu} translations, the average scores  range from $2.404$ to $3.053$, and for the low-\textsc{bleu} subset, they range from $1.281$ to $1.842$. As the LLM judge assigns lower scores to the low-\textsc{bleu} subset across all criteria than the human evaluator, it may rely more strongly on lexical similarity signals related to \textsc{bleu}, while humans evaluate quality criteria more independently.

Correlation analysis between human and LLM individual scores shows moderate to strong agreement for \textsl{completeness} ($\rho = 0.766$) and \textsl{structure} ($\rho = 0.582$). \textsl{Clarity} ($\rho = 0.196$) and \textsl{correctness} ($\rho = 0.141$), in turn, show weak correlations, which suggest that the LLM judge does not reliably reproduce human judgments for fluency-related aspects of translation.

\subsection{Human report on translation errors}
The error analysis of Polish translations generated by IPIS-tuned Bielik-11B reveals several reoccurring errors affecting \textsl{completeness}, \textsl{clarity}, \textsl{structure} and linguistic \textsl{correctness}.

\paragraph{Completeness.} Many translations contain hallucinated content, where Bielik-11B introduces information that is not present in the English source text or in the Polish reference. Furthermore, some passages from the original text are omitted, and the translations are thus incomplete. Another common issue is the repetition or paraphrasing of the same content across multiple items in a list.

\vspace*{-.2em}
\paragraph{Clarity.} Problems with clarity and the logical structure are often caused by weak connections between passages in the translated text. In some cases, sentences addressing a~similar topic appear detached from their original context, which makes the translation difficult to follow.

\vspace*{-.2em}
\paragraph{Structure.} Structural errors occur when Bielik-11B hallucinates new items (e.g. numbered points) or omits existing ones. This results in translations that do not faithfully reproduce the structure of the source text.

\vspace*{-.2em}
\paragraph{Linguistic correctness.} Translations also contain grammatical and typographical errors, e.g. incorrect  or non-agreeing verb forms (Ex. \ref{item:1}), missing prepositions in prepositional phrases (Ex. \ref{item:2}), or inconsistent font usage (Ex. \ref{item:3}). Examples:

\begin{enumerate}
    \item \label{item:1} \parbox[t]{\linewidth}{\textit{\underline{Członk$*$owie/inie} Komisji powoływan$*$i/e między datą wejścia w życie Traktatu z Lizbony a 31 października 2014 roku \underline{wybiera się}} [\textcolor{Green}{wybieran$*$i/e są}] \textit{spośród obywatel$*$i/ek Państw Członkowskich.}\\[0.3ex][{\fontsize{10}{10}\selectfont Eng. \underline{Members$_{\text{masc}/\text{fem}}$} of the Commission appointed$_{\text{masc}/\text{fem}}$ between the date of entry into force of the Treaty of Lisbon and 31 October 2014 \underline{are selected$_{\text{masc}/\text{fem}}$} from among the citizens$_{\text{masc}/\text{fem}}$ of the Member States.}]}
    
    \item \label{item:2} \parbox[t]{\linewidth}{\textit{Rada składa się z przedstawiciel$*$i/ek rządów Państw Członkowskich, \underline{szczeblu ministerialnym}} [\textcolor{Green}{na} szczeblu ministerialnym]\textit{, któr$*$zy/e mogą...} \\[0.3ex]
    [{\fontsize{10}{10}\selectfont Eng. The Council consists of representatives$_{\text{masc}/\text{fem}}$ of the governments of the Member States \underline{at} ministerial level who$_{\text{masc}/\text{fem}}$ may...}]}
    
    \item \label{item:3} \parbox[t]{\linewidth}{\textit{\underline{PRAGNĄc} pogłębić...}\\[0.3ex]
    [{\fontsize{10}{10}\selectfont Eng. \underline{DESIRIng} to deepen...}]}
    
\end{enumerate}

\subsection{Evaluation of gender-inclusive modifications}
\label{sec:apx:modifications}

\begin{table*}[t]
    \renewcommand\tabcolsep{7pt}
    \renewcommand*{\arraystretch}{1.1}
    \small
\begin{tabular}{l|l||rrrr|r}%
    \toprule
        \multirow{2}{*}{\textbf{Test Split}} & \multirow{2}{*}{\textbf{Evaluator}} & \multicolumn{4}{c|}{\textbf{MT-related criteria}} & \multirow{2}{*}{\textbf{Gender inclusivity}}\\
         & & \textbf{Completeness} & \textbf{Clarity} & \textbf{Structure} & \textbf{Correctness} & \\

\midrule
\multirow{2}{*}{\textbf{high-\textsc{bleu}}}  & LLM & 2.632 & 2.614 & {3.053} & 2.404 & {3.263} \\
 & Human & {4.053} & {4.947} & {4.632} & {4.965} & {4.246} \\
 \midrule
\multirow{2}{*}{\textbf{low-\textsc{bleu}}}  & LLM & {1.281} & {1.491} & {1.281} & {1.842} & {2.947} \\
 & Human & {1.333} & {4.333} & {3.018} & {4.509} & {3.579} \\
 \midrule
 \midrule
 \multicolumn{7}{c}{\textbf{Pearson and Spearman correlations}}\\
 \midrule
 \multicolumn{2}{l|}{\textbf{Pearson r} (p-value)} & 0.681 (0.000) & 0.145 (0.123) & 0.471 (0.000) & 0.154 (0.102) & 0.517 (0.000) \\
 \multicolumn{2}{l|}{\textbf{Spearman  $\rho$} (p-value)} & 0.766 (0.000) & 0.196 (0.037) & 0.582 (0.000) & 0.141 (0.133) & 0.539 (0.000) \\
 \bottomrule
 \end{tabular}
\caption{Comparison of LLM-as-a-judge and human evaluations of English-to-Polish test translations generated by IPIS-tuned Bielik-11B (without a system prompt). The generated translations are split into the high-\textsc{bleu} and the low-\textsc{bleu}-scored subsets (57 translations each). The LLM-judge is Qwen3-30B-A3B-Thinking-2507. The upper part of the table reports average scores; the bottom part presents the correlation between human and LLM scores across all evaluation criteria and test instances.}
\label{tab:llm-as-a-judge-table}
\vspace*{-.3em}
\end{table*}

Assessing English$\rightarrow$Polish translations, the human evaluator counts gender-inclusive modifications (Table \ref{tab:statistics_modifications}). In the high-\textsc{bleu} subset, $78.7\%$ of candidates (i.e. words requiring inclusive modification) contain gender-inclusive forms, compared to $58.6\%$ in the low-\textsc{bleu} subset. Many of them are correctly introduced gender-inclusive expressions ($62.8\%$ of the high-\textsc{bleu} instances, and $48.3\%$ of the low-\textsc{bleu} ones).

\begin{table}[!ht]
    \centering
    \renewcommand\tabcolsep{4.5pt}
    \renewcommand*{\arraystretch}{1.1}
    \small
    \begin{tabular}{l|rr|rr}
    \toprule
       \multirow{2}{*}{\textbf{Category}} & \multicolumn{2}{c|}{\textbf{high-\textsc{bleu}}} & \multicolumn{2}{c}{\textbf{low-\textsc{bleu}}}\\
       & $\#$ & \% & $\#$ & \% \\
       \midrule
        Candidates for incl. changes &  239 & &  273 & \\
        \midrule
        Modified candidates: & 188 & 78.7 & 160 & 58.6\\
        $\rightarrow$ correct incl. forms & 150 & 62.8 & 132 & 48.3\\
        $\rightarrow$ incorrect incl. forms & 38 & 11.7 & 28 & 10.3\\
        \midrule
        Missing incl. changes & 51 & 21.3 & 113 & 41.4\\
        \midrule
        Non-personal changes & 1 & n/a & 1 & n/a \\
        \bottomrule
    \end{tabular}
    \caption{Statistical overview of gender-inclusive expressions in the test translations generated by IPIS-tuned Bielik-11B.}
    \label{tab:statistics_modifications}
\vspace*{-.3em}
\end{table}

The rate of incorrect inclusive forms is relatively similar across all subsets. Incorrect inclusive forms occur in $11.7\%$ of high-\textsc{bleu} translations and $10.3\%$ of low-\textsc{bleu} ones. This suggests that translation quality (as measured by \textsc{bleu}) does not strongly affect the likelihood of producing incorrect inclusive forms. Examples of incorrect gender-inclusive expressions:
    \begin{itemize}
        \item \parbox[t]{\linewidth}{[incorr.] \textit{\underline{wysoki*a} przedstawiciel*ka jest \underline{wspomagany*a}...}\footnote{The underlined expressions could be valid gender-inclusive forms. However, as the models were prompted to produce only the variant of inclusive abbreviated forms proposed by \citet{wro:inni:2025}, these inconsistent generations are counted as incorrect.}\\[0.3ex]
        [corr.] \textit{\textcolor{Green}{wysok*i/a} przedstawiciel*ka jest \textcolor{Green}{wspomagan*y/a}...}\\[0.3ex]
    {\fontsize{10}{10}\selectfont [Eng.] the High$_{\text{masc}/\text{fem}}$ Representative$_{\text{masc}/\text{fem}}$ shall be assisted$_{\text{masc}/\text{fem}}$...}}
        \item \parbox[t]{\linewidth}{[incorr.] \textit{mieszka\underline{n}*com/kom}\\[0.3ex]
        [corr.] \textit{mieszka*\textcolor{Green}{ń}com/\textcolor{Green}{n}kom} \\[0.3ex]
    {\fontsize{10}{10}\selectfont [Eng.] residents$_{\text{masc}/\text{fem}}$}}
        \item \parbox[t]{\linewidth}{[incorr.] \textit{szefo*wie/\underline{wo}we}\\[0.3ex]
        [corr.] \textit{szefo*wie/we}\\[0.3ex]
        {\fontsize{10}{10}\selectfont [Eng.] bosses$_{\text{masc}/\text{fem}}$}}
        
    \end{itemize}

\noindent
$41.4\%$ of low-\textsc{bleu} candidate translations lack the required gender-inclusive expressions, compared to only $21.3\%$ in the high-\textsc{bleu} subset. A major reason for this is paraphrasing. IPIS-tuned Bielik-11B sometimes generates several alternative versions of a source text passage, some of which include gender-inclusive forms while others do not. This does not necessarily reflect a weakness of the model; rather, it suggests that the model is uncertain about which variant is most appropriate.

As only one instance in each subset contains an incorrect modification of a non-personal word, such errors appear to be extremely rare in translations generated by IPIS-tuned Bielik-11B, e.g.:

\begin{itemize}
        \item \parbox[t]{\linewidth}{[incorr.] \textit{\underline{Żadn$*$emu/ej} \textit{Państwu Członkowskiemu}}\\[0.3ex]
        [corr.] \textcolor{Green}{Żadnemu} Państwu Członkowskiemu\\[0.3ex]
        {\fontsize{10}{10}\selectfont [Eng.] No Member State}}
        \item \parbox[t]{\linewidth}{[incorr.] \textit{Parlament Europejski wybiera swoj*ego/ą przewodnicząc*ego/ą i \underline{sw*oj/e}} \textit{organy...} \\[0.3ex]
        [corr.] Parlament Europejski wybiera swoj*ego/ą przewodnicząc*ego/ą i \textcolor{Green}{swoje} organy... \\[0.3ex]
        {\fontsize{10}{10}\selectfont [Eng.] The European Parliament elects its$_{\text{masc}/\text{fem}}$ President$_{\text{masc}/\text{fem}}$ and \underline{its} bodies...}}
        
\end{itemize}

\end{document}